\documentclass[runningheads]{llncs}
\usepackage{graphicx}
\usepackage{comment}
\usepackage{amsmath,amssymb} 
\usepackage{color}
\usepackage{multirow}
\usepackage{wrapfig,lipsum,booktabs}
\usepackage{pifont}
\usepackage{appendix}
\usepackage[pagebackref=true,breaklinks=true,letterpaper=true,colorlinks,bookmarks=false]{hyperref}

\newcommand{\cmark}{\text{\ding{51}}}
\newcommand{\xmark}{\text{\ding{55}}}
\newcommand{\eg}{\textit{e.g.}}
\newcommand{\ie}{\textit{i.e.}}
\newcommand{\etc}{\textit{etc.}}
\newcommand{\etal}{\textit{et al.}}

\begin{document}
\pagestyle{headings}
\mainmatter

\def\ACCV20SubNumber{803}  

\title{A Benchmark and Baseline for\\ Language-Driven Image Editing} 
\titlerunning{A Benchmark and Baseline for Language-Driven Image Editing}
\authorrunning{J. Shi et al.}

\author{Jing Shi$^{1}$, Ning Xu$^2$,  Trung Bui$^2$,  Franck Dernoncourt$^2$, Zheng Wen$^{2}$, and Chenliang Xu$^1$\\
$^1$University of Rochester \quad\quad $^2$Adobe Research\\
$^1${\tt\small \{j.shi,chenliang.xu\}@rochester.edu} \\
$^2${\tt\small \{nxu,bui,dernonco\}@adobe.com zhengwen@alumni.stanford.edu}}
\institute{}

\maketitle

\begin{abstract}
Language-driven image editing can significantly save the laborious image editing work and be friendly to the photography novice.
However, most similar work can only deal with a specific image domain or can only do global retouching. 
To solve this new task, we first present a new language-driven image editing dataset that supports both local and global editing with editing operation and mask annotations.
Besides, we also propose a baseline method that fully utilizes the annotation to solve this problem. 
Our new method treats each editing operation as a submodule and can automatically predict operation parameters. Not only performing well on challenging user data, but such an approach is also highly interpretable. We believe our work, including both the benchmark and the baseline, will advance the image editing area towards a more general and free-form level.
\end{abstract}

\section{Introduction}
There are numerous reasons that people want to edit their photos,~\eg, remove tourists from wedding photos, improve saturation and contrast to make photos more beautiful, or replace background simply for fun. Therefore, image editing is very useful and important in people's everyday life. However, it is not a simple task for most people. One reason is that current mainstream photo editing softwares (\eg~Photoshop) could work only if users understand the concept of various editing operations such as hue, saturation, selection~\etc, and know how to use them step by step. However, most novice users do not have such knowledge. Another reason is that most editing operations require some manual work, some of which could be very time-consuming. It is even more challenging when editing photos on mobile devices because people have to use their fingers while screen sizes are small.

\begin{figure}[t]
    \centering
    \includegraphics[width=\columnwidth]{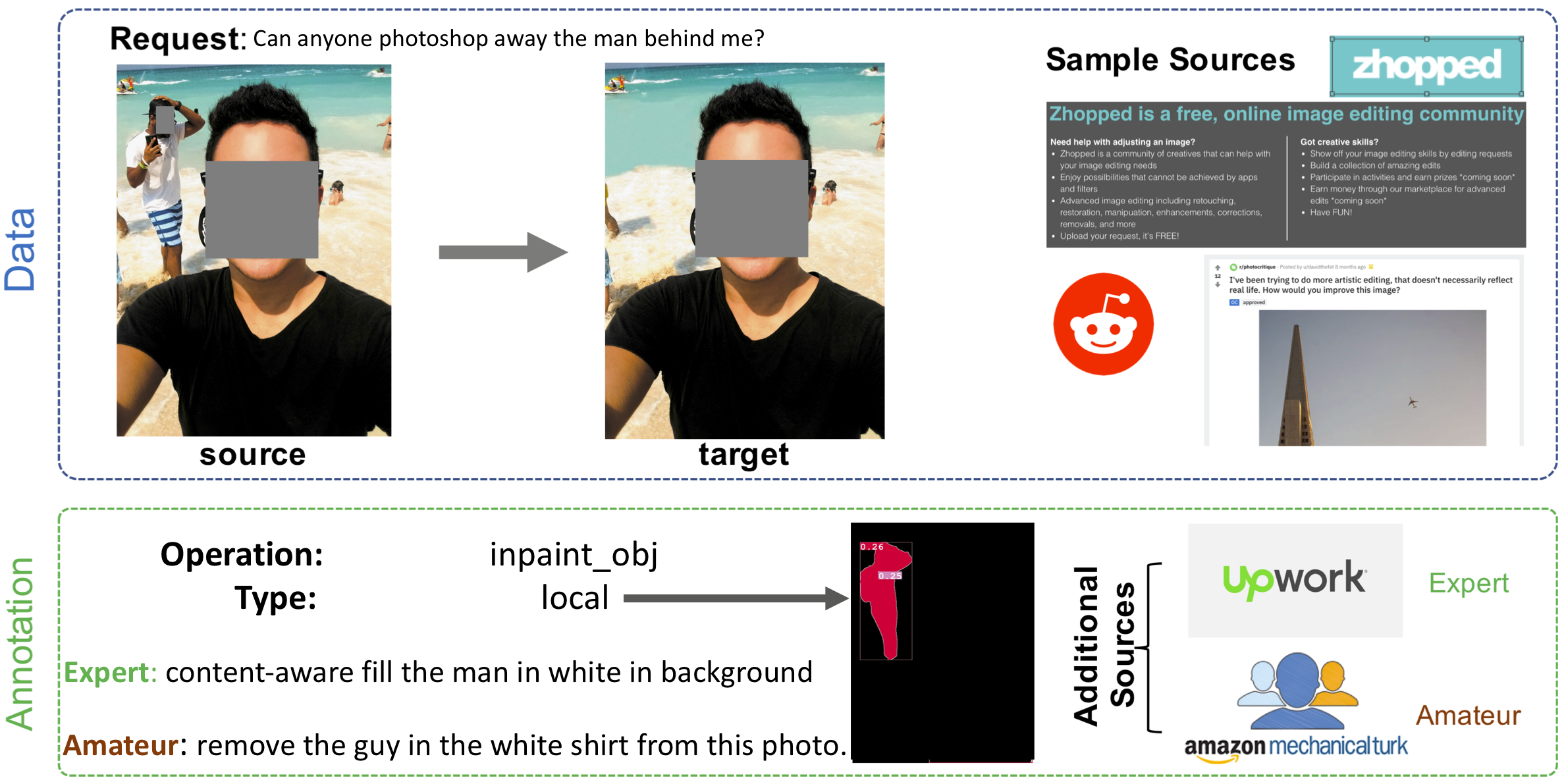}
    \caption{One example in our newly collected Grounded Image Editing Request (GIER) dataset. Each sample is a triplet. We collected all samples from image-editing-request websites, \eg, Zhopped and Reddit, and we augment language request data from both experts (Upwork) and the crowd-sourcing website (AMT).} 
    \label{fig:data_indication}
\end{figure}

\newcommand{\ra}[1]{\renewcommand{\arraystretch}{#1}}
\begin{table*}[h]
    \caption{Comparison between GIER dataset and related existing datasets. \textit{Size} is the number for unique images or image pairs (if paired). \textit{User photo} and \textit{user request} mean the image and request are general and are from real user. \textit{Other annotation} is the annotation of editing mask or editing operation.}
    \centering
    \ra{1.2}
    \scalebox{0.9}{
    \begin{tabular}{cccccc}
    \toprule
    \textbf{dataset} & \textbf{size} & \textbf{user photo} & \textbf{user request} & \textbf{paired image} & \textbf{other annotations} \\
    \midrule
    CUB~\cite{wah2011caltech} & 11788 &\xmark  &\xmark  &\xmark &\xmark \\
    Oxford-102~\cite{nilsback2008automated}  & 8189 &\xmark  &\xmark  &\xmark &\xmark \\
    DeepFashion-Seq~\cite{zhu2017your} & 4820 &\xmark  &\xmark  &\cmark  &\xmark\\
    CoDraw~\cite{el2019tell} & 9993 &\xmark  &\xmark  &\cmark &\xmark \\
    i-CLEVER~\cite{el2019tell} & 10000 &\xmark  &\xmark  &\cmark &\xmark\\
    \textbf{IGER~(Ours)} & 6179 &\cmark &\cmark  &\cmark  &\cmark\\
    \bottomrule
    \end{tabular}}
    \label{tab:dataset}
 \end{table*}

In this paper, we propose \textit{language-driven image editing}~(LDIE) to make image editing easier for everybody. Specifically, to edit an image, a user only needs to provide a natural language request. Our new algorithm will automatically perform all the editing operations to produce the desired image without any manual intervention. The language request can be very detailed, including step-wise instructions such as ``increase the brightness and reduce the contrast.'' But, it could also contain a certain level of vagueness (\eg~``make the sky bluer") or even very vague descriptions like ``make the image more beautiful," which is particularly useful for novice users. One considerable advantage of the new task is that users no longer need to be involved in the tedious editing process (\eg,~determine editing operations and sequences, manual adjustment of parameters, masking~\etc), which can be all accomplished by algorithms automatically.

There are a few previous studies that work on similar problems, but none of them can solve our new task. Many works~\cite{el2019tell,nam2018text,cheng2018sequential,shinagawa2018interactive} explore language-based manipulation for simple image contents such as birds, faces, or shoes. \cite{hu2018exposure,park2018distort} only handle the image retouching operation and do not take any language inputs. Although being language-based,~\cite{wang2018learning,chen2018language} only solve a single task in image editing (\eg,~retouching~\cite{wang2018learning} or recoloring~\cite{chen2018language}), which are not extendable to other operations. PixelTone~\cite{laput2013pixeltone} solves the problem most similar to ours. However, it requires users to select editing regions manually and can only work for very detailed instructions.

Since no previous works directly solve our new task, we tackle it in two steps. We first collect a dataset named Grounded Image Editing Request (GIER) with 30k samples and 6k unique images, where each sample is a triplet, including one source image, one language request to edit the source image, and one target image which matches the editing request. Table.~\ref{tab:dataset} illustrates the comparison of our datasets against the previous one and reflects the advantages of ours. All our image samples are real data collected from image-editing-request websites Zhopped.com and Reddit.com. We also augment language request data from both the crowd-sourcing website (AMT) and contracted experts (Upwork). We believe our dataset will become an important benchmark in this domain, given its scale and high-quality annotation. 

Next, we propose a baseline algorithm for the new task. Given a source image and a language request, our algorithm first predicts desired editing operations and image regions associated with each operation. Then, a modular network that comprises submodules of the predicted operations is automatically constructed to perform the actual editing work and produce the output image. The parameters of each operation are also automatically determined by the modular network. One advantage of our algorithm is its interpretability. At every step, it will produce some human-understandable outputs such as operations and parameters, which can be easily modified by users to improve the editing results. Besides, our method fully leverages all the dataset annotation, and each of its components helps check the quality of the dataset annotation. 

We train our algorithm on our newly collected GIER dataset and evaluate it with ablation studies. Experimental results demonstrate the effectiveness of each component of our method. Thus, our method also sets a strong baseline result for this new task. 

In summary, the contributions of this paper include:
\begin{itemize}
    \item We propose a new LDIE task that handles both detailed and vague requests on natural images as well as both global and local editing operations.
    \item We collect the first large-scale dataset, which comprises all real user requests and images with high-quality annotations.
    \item We propose a baseline algorithm that is highly interpretable and works well on challenging user data.
\end{itemize}

The rest of the paper is organized as follows. In Sec.~\ref{sec:related_work}, we briefly introduce related work. 
Section.~\ref{sec:dataset} describes our dataset and Sec.~\ref{sec:methodology} describes the proposed algorithm in detail. Experimental results are given in Sec.~\ref{sec:experiment}, and finally, we conclude the paper in Sec.~\ref{sec:conclusion}.

\section{Related Work}
\label{sec:related_work}

\noindent \textbf{Language-based image manipulation/generation.} \quad 
Many methods have been proposed recently for language-based image generation~\cite{reed2016generative,xu2018attngan,zhang2017stackgan} and manipulation~\cite{el2019tell,nam2018text,cheng2018sequential,shinagawa2018interactive,chen2018language}. In this paper, we propose a new task, LDIE, which automatically performs a series of image editing operations given natural language requests. Methods for image manipulation/generation are dominated by variants of generative adversarial networks (GAN), and they change specific attributes of simple images, which usually only contain one primary object, e.g., faces, birds, or flowers. In contrast, our new task works on everyday image editing operations (\eg,~contrast, hue, inpainting \etc) applied to more complex user images from open-domain. A recent work~\cite{wang2018learning} was proposed to edit images with language descriptions globally, and it collected a dataset that contains only global image retouching operations. In contrast, our new task handles both global and local editings, and our dataset comprises all real user requests, which cover diverse editing operations. 

\noindent \textbf{Image Editing.} \quad 
The task of image editing involves many subtasks such as object selection, denoising, shadow removal~\etc~Although many methods have been proposed for each subtask, how to combine the different methods of different subtasks to handle the more general image editing problem was seldom studied before. Laput~\etal~\cite{laput2013pixeltone} proposed a rule-based approach which maps user phrases to specific image editing operations, and thus does not require any learning. However, this method is quite limited in that it requires each request sentence to describe exactly one editing operation. Besides, it cannot automatically determine operation parameters. There are several works~\cite{hu2018exposure,park2018distort,wang2018learning} proposed for image retouching, which consider multiple global editing operations such as brightness, contrast, and hue. In~\cite{hu2018exposure,park2018distort}, reinforcement learning is applied to learn the optimal parameters for each predefined operation. \cite{wang2018learning} leverages GANs to learn each operator as a convolutional kernel. Different from these two types of methods, we employ a modular network that was previously proposed for VQA~\cite{andreas2016neural,hu2017learning,mao2019neuro}, and learn optimal parameters for each predefined operator in a fully supervised manner. 

\noindent \textbf{Visual Grounding.} \quad 
Our algorithm needs to decide whether an operator is applied locally or globally and where the local area is if it is a local operator. Therefore, visual grounding is an essential component of our algorithm. However, previous visual grounding methods~\cite{kazemzadeh2014referitgame,hu2016natural,wang2018learning,yang2019fast,liu2019improving} are not directly applicable to our task due to the complexity of our language requests. For example, an expression for traditional visual grounding only contains the description of a single object. In contrast, our request contains not only object expression but also other information such as editing operators. Besides, each request may include multiple operators, and each could be a local one. Furthermore, each local region is not necessarily a single object. It could also be a group of objects or stuff (\eg,~``remove the five people on the grass''). Therefore the visual grounding problem is more challenging in our task.

\begin{wraptable}{r}{0.45\textwidth}
\vspace{-11mm}
    \centering
    \caption{Statistics of all candidate operations. Each column represents the operation, the number of occurrence for each operation, the ratio of each operation over all operations , the ratio of images containing the operation over all images, and the ratio of local operation for each operation.}
    \label{tab:op_ratio}
    \scalebox{0.82}{
    \begin{tabular}{@{}lrrrr@{}}
    \toprule
    operation         & \#occur & opr\% & img\% & local\%\\ \midrule
    brightness        & 3176      & 16.00           & 51.40       & 7.53        \\
    contrast          & 3118      & 15.70           & 50.46       & 4.84        \\
    saturation        & 2812      & 14.16           & 45.51       & 7.15        \\
    lightness         & 2164      & 10.90           & 35.02       & 6.93        \\
    hue               & 2059      & 10.37           & 33.32       & 11.56       \\
    remove object     & 1937      & 9.76            & 31.35       & 99.59       \\
    tint              & 1832      & 9.23            & 29.65       & 7.59        \\
    sharpen           & 842       & 4.24            & 13.63       & 6.18        \\
    remove bg         & 495       & 2.49            & 8.01        & 95.35       \\
    crop              & 405       & 2.04            & 6.55        & 23.95       \\
    deform object     & 227       & 1.14            & 3.67        & 17.18       \\
    de-noise          & 155       & 0.78            & 2.51        & 9.68        \\
    dehaze            & 133       & 0.67            & 2.15        & 11.28       \\
    gaussain blur     & 124       & 0.62            & 2.01        & 73.39       \\
    exposure          & 85        & 0.43            & 1.38        & 5.88        \\
    rotate            & 84        & 0.42            & 1.36        & 1.19        \\
    black\&white      & 72        & 0.36            & 1.17        & 16.67       \\
    radial blur       & 65        & 0.33            & 1.05        & 83.08       \\
    flip image        & 23        & 0.12            & 0.37        & 0.00        \\
    facet filter      & 19        & 0.10            & 0.31        & 5.26        \\
    rotate object     & 12        & 0.06            & 0.19        & 41.67       \\
    find edges filter & 10        & 0.05            & 0.16        & 0.00        \\
    flip object       & 7         & 0.04            & 0.11        & 85.71       \\
    \bottomrule
    \end{tabular}}
    \vspace{-13mm}
\end{wraptable}

\section{The Grounded Image Editing Request (GIER) Dataset}
\label{sec:dataset}
In this section, we present how we collect a large-scale dataset called \textit{Grounded Image Editing Request} (GIER) to support our new task.

\subsection{Dataset Collection}
\noindent \textbf{Step 1: Preparation.} \quad 
First, we crawl user data from two image editing websites: Zhopped\footnote{\url{http://zhopped.com}} and Reddit\footnote{\url{https://www.reddit.com/r/photoshoprequest}}. On the websites, amateur photographers post their images with editing requests, which are answered by Photoshop experts with edited images. Our crawled web data spans from the beginning of the two websites until 4/30/2019, resulting in 38k image pairs. Then, we construct a list of editing operations that cover the majority of the editing operations of the crawled data, which is shown in Tab.~\ref{tab:op_ratio}. 

\noindent \textbf{Step 2: Filtering and Operation Annotation.} \quad 
Although the crawled dataset is large, many samples are too challenging to include. There are mainly two challenges. First, some images contain local editing areas, which are hard to be grounded by the existing segmentation models due to the lack of training labels or other reasons. Second, some editing requests involve adding new objects or background into the original images, which cannot be easily handled by automatic methods.

To make our dataset more practical, we ask annotators to filter crawled samples belonging to the two challenging cases. To decide whether a local editing operation can be grounded or not, we preprocess each original image by applying the off-the-shelf panoptic segmentation model UPSNet~\cite{kirillov2019panoptic} and let annotators check whether the edited areas belong to any pre-segmented regions.

After the filtering work, annotators are further asked to annotate the qualified samples with all possible operations from the operation list in Tab.~\ref{tab:op_ratio} as well as the edited region of each operator. To get better-quality annotation, we hire Photoshop experts from Upwork to do the tasks. After the first-round annotation, we do another round of quality control to clean the low-quality annotations.

\noindent \textbf{Step 3: Language Request Annotation.} \quad
The crawled web data already contains one language request per sample. However, sometimes the original requests do not match the edited images well, which could cause problems for model training. Besides, we are interested in collecting diverse requests from different levels of photographers. Therefore we collect language requests from both the AMT and Upwork. AMT annotators usually have less knowledge about image editing, and Upwork annotators are Photoshop experts. 

We present pairs of an original image and an edited image to AMT annotators without showing anything else. This is different from what we give to Upwork annotators, which contains additional information, including the original request as well as the annotation in step 2. To balance the data distribution, we collect three requests from AMT and two requests from Upwork. We also do another round of quality control to clean the bad annotations.

\subsection{Data Statistics}
The GIER dataset contains 6,179 unique image pairs. Each pair is annotated with five language requests, all possible editing operations, as well as their masks. 
The average number of operations per image pair is 3.21, and the maximum is 10. The distribution of each operation is show in Tab.~\ref{tab:op_ratio}. For language requests, the average word length is 8.61; the vocabulary size is 2,275 (after post-processing). 

Our newly-collected GIER dataset is highly valuable for LDIE task. First, all data are from real users' editing requests so that they genuinely reflect a large portion of the needs for image editing. Second, we collect language requests from diverse users, which are helpful in making methods trained on our dataset practical for real applications.
Third, our dataset is annotated with many different types of ground truth, which makes learning different types of methods possible.

\begin{figure}[t] 
\centering\includegraphics[width=\columnwidth]{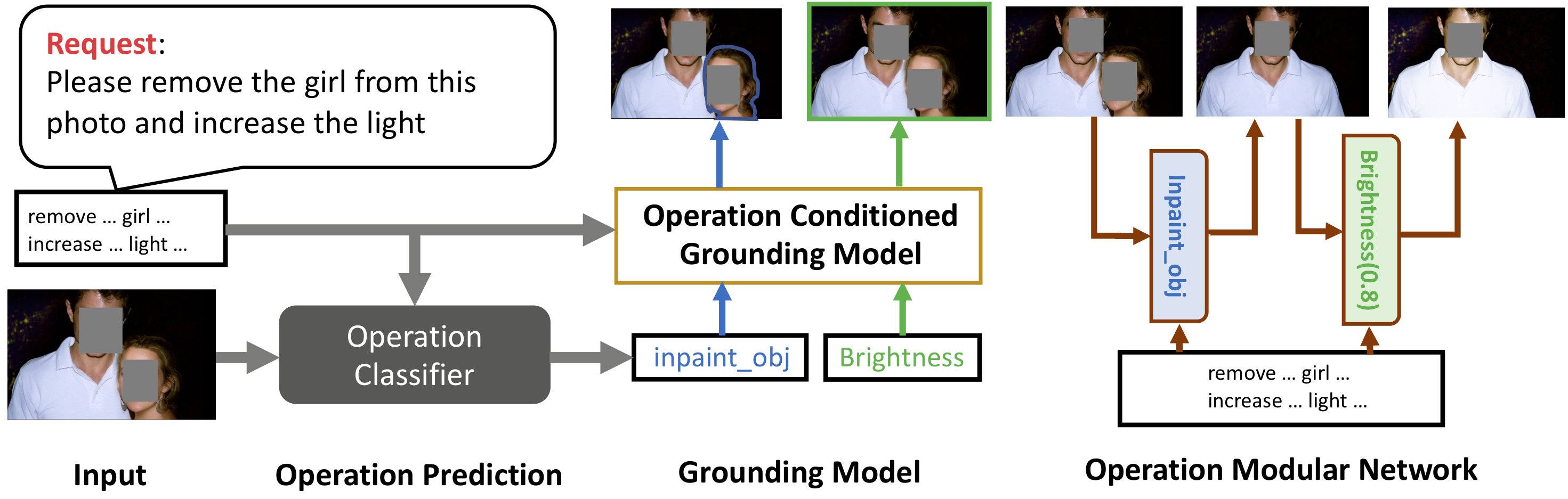}
\caption{An overview of the model's pipeline. The input image and request are sent to a multi-label classifier to predict operations. Then, the operation conditioned grounding model takes in the image, the request, and operations and outputs the grounding mask for each operation. Finally, the operation modules are cascaded to form the operation modular network, and each step outputs an interpretable intermediate result.}
\label{fig:pipeline}
\end{figure}

\section{A Baseline for Language-Driven Image Editing}
\label{sec:methodology}
    
We define the task of LDIE as follows. Given an original image and a user request, a method needs to produce an output image that matches the editing request. The closer the output image is to the target image, the better the method is. 
Contrast from the prevalent GAN-based methods, we propose the baseline model that can edit by sequentially applying interpretable editing operations, requiring the comprehension of both language request and visual context. 
Since most operations are resolution-independent, our model can also keep the image resolution same as the input.
The main body of our model is an operation modular network (Sec.~\ref{sec:OpModualrNet}) shown in our model pipeline (Fig.~\ref{fig:pipeline}). 
It stacks multiple editing operations in order and predicts best parameters.
Since the layout of operations is discrete variable which is hard to optimize only given the target image, we resort to a supervisely trained operation classifier (Sec.~\ref{sec:OpPred}) to predict needed editing operations and arrange them in a fixed order.
Moreover, every editing operation requires a mask to specify where to edit, which is obtained by the operation conditioned grounding model (Sec.~\ref{sec:OpCondGround}).
Although our model is not completely end-to-end trained, it is a valuable initial attempt to address such task in such a compositional and interpretable way.

\subsection{Operation Prediction} 
\label{sec:OpPred}

Since samples in GIER are annotated with ground truth operations, we train a multi-label classification network to predict operations. In Tab.~\ref{tab:op_ratio}, there are 23 operations, while some of them have too few training examples. Therefore, we pick the top nine operations ($\mathtt{brightness}$ and $\mathtt{lightness}$ are merged as one operation due to their similarity) as our final classification labels. They cover 90.36\% of total operations and are representative of users' editing habits. Both input image and language request are input to the classifier, owning to many unspecific requests which require the perception of the input image. The image is embedded by ResNet18~\cite{he2016deep}, and the language request is embedded by a bi-directional LSTM~\cite{schuster1997bidirectional}. The two features are then concatenated and passed into several fully connected layers to get the final prediction. This model is trained with the multi-label cross-entropy loss. 

\begin{figure}[t]
    \centering
    \includegraphics[width=\columnwidth]{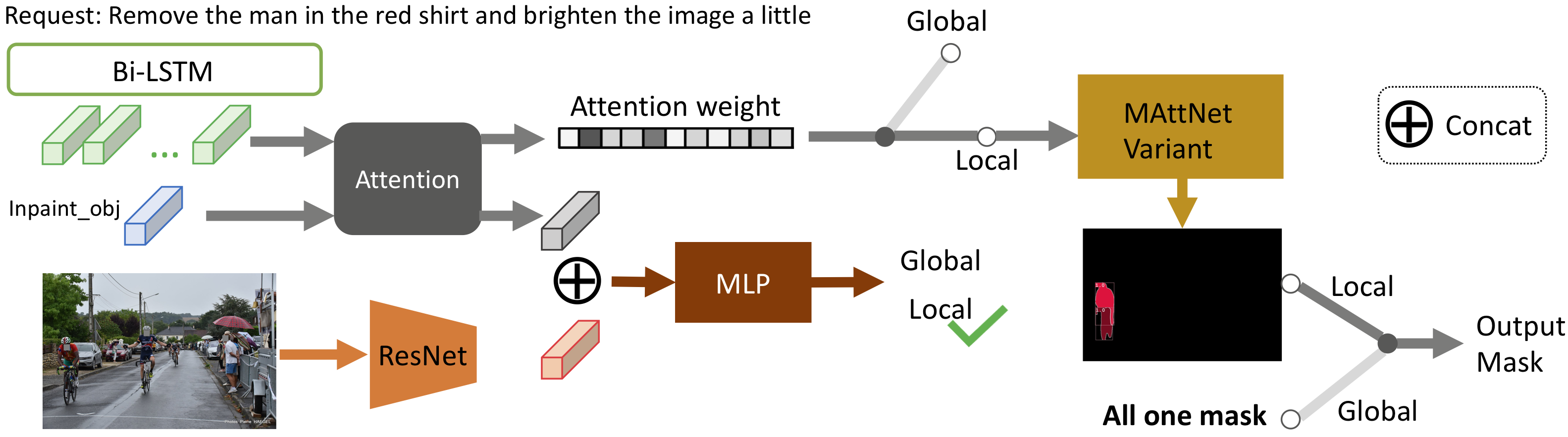}
    \caption{Network structure of the operation-conditioned grounding model. The operation attends to its related part in the request. And the MLP binary classifier can judge if the operation is local or global. If local, the MAttNet variant will ground the operation related description into mask, otherwise output the all one mask.}
    \label{fig:opCondGround}
\end{figure}

\subsection{Operation Conditioned Grounding Model} \label{sec:OpCondGround}

In our task, the language of request may contain multiple types of operation-based groundings (\eg, ``please remove the girl from this photo and increase the light'' in Fig.~\ref{fig:pipeline}) and each grounding may contain multiple, even disconnected regions (\eg, ``remove all pedestrians from the image''). Given such uniqueness of our task, the previous visual grounding methods are not directly applicable. However, they certainly serve as a good starting point. In this section, we will first review a state-of-the-art visual grounding model, MattNet~\cite{yu2018mattnet}, and then show step-by-step how to turn it into a proper grounding model for its use in our task taking into consideration of the operation input and multi-region output.


\noindent\textbf{The Grounding Problem and MattNet.} \quad 
Given a request $Q$, an operation $O$, and an image $I$, the goal is to localize the area of the image to be edited by the operator. We formulate it as a retrieving problem by extracting the region proposals $R=\{R_i\}$ from the image and choosing one or more of the region proposals to make up the target area.

The basic MattNet comprises a language attention network and three visual modules---subject, location, and relationship. The language attention network takes the query $Q$ as input and outputs the modular phrase embeddings for each module $[q^{subj} , q^{loc}, q^{rel}]$ and their attention weights $\{w_{subj} , w_{loc}, w_{rel}\}$. Each module individually calculates the matching score between its query embedding and the corresponding visual feature. Then the matching scores from three modules are weighted averaged by $\{w_{subj} , w_{loc}, w_{rel}\}$. Finally, the ranking loss for positive query-region pair $(Q_i, R_i)$ and negative pair $(Q_i, R_j)$ is: 
\begin{align}
    L_{rank} = \Sigma_i(&\max(0, \Delta + s(Q_i, R_j) - s(Q_i, R_i)) + \nonumber \\
    &\max(0, \Delta + s(Q_j, R_i) - s(Q_i, R_i))),
    \label{eq:ranking}
\end{align}
where $s(x, y)$ denotes the matching score between query $x$ and region $y$, and $\Delta$ denotes the positive margin. 

\noindent\textbf{Operation Conditioned Language Attention.} \quad 
We extend the language attention network of MattNet. The reason for choosing MattNet is that the editing request frequently describes the objects in the subject-location-relationship format. The request $Q$ of length $T$ represented by word vectors $\{e_t\}_{t=1}^T$ is encoded by a Bi-LSTM and yields the hidden vectors $\{h_t\}_{t=1}^T$. The operation word embedding is $o$. The operation finds its corresponding noun phrase in the request by using attention. Therefore, the attention weights from the operation to all the request tokens are: 
\begin{equation} \label{eq:op_attn}
    \alpha_t^{(o)} = \frac{\exp(\langle{o},h_t\rangle)}{\sum_{k=1}^T\exp(\langle{o},h_k\rangle)},
\end{equation}
where $\langle \cdot, \cdot\rangle$ denotes inner product, and the superscript $(o)$ indicates the specific attention for operation $o$.  
We keep trainable vectors $f_m$, where $m\in\{\text{subj, loc, rel}\}$, from MattNet to compute the attention weights for each of three visual modules: 
\begin{equation}
    a_{m,t} = \frac{\exp(\langle{f_m},h_t\rangle)}{\sum_{k=1}^T\exp(\langle{f_m},h_k\rangle)}.
\end{equation}
Then, we can compute an operation conditioned attention and thus, obtain operation conditioned modular phrase embedding: 
\begin{equation}
    \hat{a}_{m,t} = \frac{\alpha_t a_{m, t}}{\sum_{k=1}^T \alpha_k a_{m, k}}, \;\;\;\;\;\;
    q_m^{(o)} = \sum_{t=1}^T\hat{a}_{m,t}^{(o)} e_t.
\end{equation}
%
%
%
The other parts of the language attention network remain the same.
For the visual modules, we keep the location and relationship modules unchanged. For the subject module, we remove the attribute prediction branch because the template parser~\cite{kazemzadeh2014referitgame} is not suitable for our editing request.

\noindent\textbf{Multiple Object Grounding.} \quad 
Since we formulate the task as a retrieving problem, we set a threshold for the matching score to determine multiple grounding objects. If all objects are under the threshold, the top-1 object will be selected. However, an operation might be grounded to the whole image, which requires the model to retrieve all the candidates. To remedy such a problem, we add an extra binary classifier to tell if the operation is local or global, given the context of image and request. The structure is presented in Fig.~\ref{fig:opCondGround}.

Since GIER dataset provides ground truth instance masks, the operation-conditioned grounding model is trained with the ranking loss as Eq.~\ref{eq:ranking}.



\subsection{Operation Modular Network} \label{sec:OpModualrNet}
\begin{figure}[t]
    \centering
    \includegraphics[width=0.7\columnwidth]{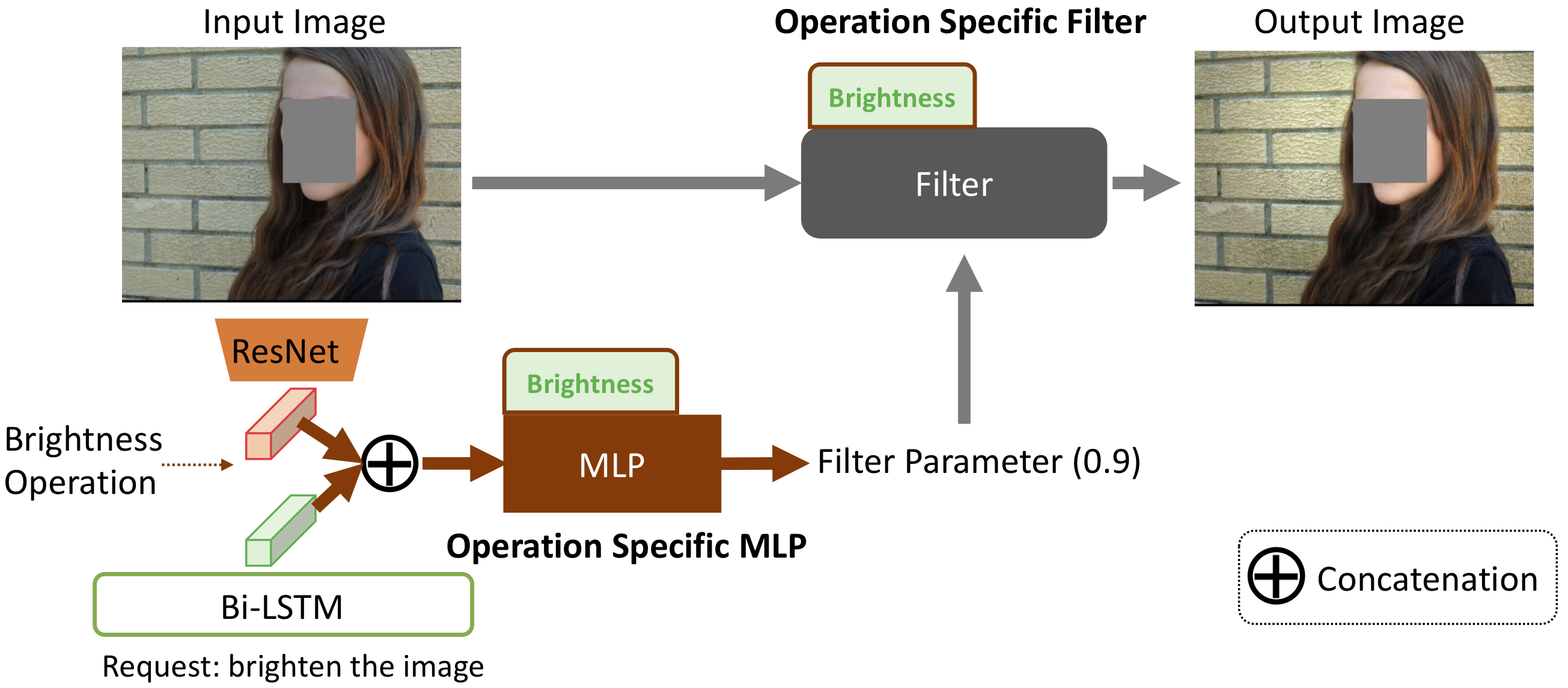}
    \caption{The structure of a submodule network for the $\mathtt{brightness}$ operation.}
    \label{fig:modular}
\end{figure}

After the set of possible operations, along with their masks, are predicted, our method constructs a Operation Modular Network~(OMN) to perform the actual editing work. The OMN is composed of submodules, each of which represents a predefined editing operation. Each submodule takes as an input image or the previously edited image, the language request and the mask, and produces an output image. See Fig.~\ref{fig:modular} for an illustration. The training objective for OMN is learning to predict the best parameter for each operation from the supervised of the target image. Next, we first describe the implementation of each submodule, then the way how we create the modular network, and finally, the loss functions. 


\noindent\textbf{Submodule Implementation.}\quad
We create one submodule for each chosen operation. Among them, six are implemented by differentiable filters which are also resolution independent. Specifically, $\mathtt{brightness}$ and $\mathtt{saturation}$ are implemented by scaling the HSV channels. $\mathtt{sharpness}$ is achieved by adding the image spatial gradient. $\mathtt{contrast}$, $\mathtt{hue}$, and $\mathtt{tint}$ are implemented the same as~\cite{hu2018exposure}. For $\mathtt{remove\_bg}$  we simply implement $\mathtt{remove\_bg}$ as changing the masked area to white given our sample distribution, which is non-differentiable. And $\mathtt{inpaint\_obj}$ is implemented by a differentiable neural inpainting model EdgeConnect~\cite{nazeri2019edgeconnect}. Refer to App.~\ref{sec:op_design} for more implementation details.

Except $\mathtt{remove\_bg}$ and $\mathtt{inpaint\_obj}$, the other operations also require some input parameters, which are automatically predicted by their submodules. Specifically, the request and image features are concatenated and sent to an MLP to predict the parameter. The filter takes in the parameter and mask and yields the output image. Each operation has its individual MLP parameters.

\noindent\textbf{Modular Network Creation.} \quad 
The modular network is created by linking together all predicted operations. However, $\mathtt{remove\_bg}$ is non-differentiable thus would blocked the gradient backpropagation. 
And $\mathtt{inpaint\_obj}$ is a large network that is computational expensive for gradient.
Luckily, these two submodules do not have any parameters to learn. Therefore, we always put them in front of the chain if they exist. 

\noindent\textbf{Loss Function.}\quad
The L1 loss is applied between the final output image $I_K$ and the target image $I_{gt}$ to drive the output image to be similar to the target image:
\begin{equation}
    \text{Loss}_{l1} = |I_K - I_{gt}|,
\end{equation}
where $K$ denotes the number of predicted operations,~\ie~the length of the submodule chain.

However, only using the supervision at the final step might not guarantee that the intermediate images are adequately learned. Hence, we also propose to use a step-wise triplet loss to enforce the intermediate image to be more similar to the target image than its previous step:
\begin{equation}
    \text{Loss}_{tri} = \frac{1}{K}\sum_{k=0}^{K-1}\max(|I_{k+1} - I_{gt}| - |I_k - I_{gt}| + \Delta, 0),
\end{equation}
where $\Delta$ is a positive margin. It resembles triplet loss by regarding $I_{gt}$ as anchor, $I_k$ as negative sample and $I_{k+1}$ as positive. Note that we should block the gradient of the term $|I_k - I_{gt}|$ to prevent from enlarging the distance between $I_{gt}$ and $I_k$. Hence final loss is $\text{Loss} = \text{Loss}_{l1} + \lambda \text{Loss}_{tri}$, with balanced weight $\lambda$.

\section{Experiment}
\label{sec:experiment}


\begin{table}[t]
    \parbox{.49\linewidth}{
        \caption{The F1 score and ROC-AUC score for operation prediction.}
        \begin{center}
        \scalebox{0.85}{
            \begin{tabular}{@{}ccccccc@{}}
            \toprule
                   & \multicolumn{5}{c}{F1}               & \multirow{2}{*}{ROC}   \\ \cmidrule{2-6} 
            threshold & 0.3   & 0.4   & 0.5   & 0.6   & 0.7   &       \\ 
            \midrule
            val    & .7658 & .7699 & .7620 & .7402 & .7026 & .9111 \\ 
            test   & .7686 & .7841 & 7759  & .7535 & .7172 & .9153 \\ 
            \bottomrule
            \end{tabular}
        }
        \end{center}
        \label{tab:multilabel_f1}
    }
    \hfill
    \parbox{.49\linewidth}{
        \caption{The operation type classification accuracy}
        \begin{center}
        \scalebox{0.85}{
            \begin{tabular}{@{}ccccccc@{}}
            \toprule
                   & \multicolumn{5}{c}{Accuracy}               &\multirow{2}{*}{ROC}   \\ \cmidrule{2-6}
            threshold & 0.1   & 0.3   & 0.5   & 0.7   & 0.9   &       \\ 
            \midrule
            val    & .9328 & .9328 & .9328 & .9328 & .9328 & .8915 \\ 
            test   & .9377 & .9387 & .9397 & .9397 & .9397 & .8969 \\ 
            \bottomrule
            \end{tabular}
        }
        \end{center}
        \label{tab:operation_type}
    }
\end{table}




\begin{table*}[t]
\caption{The grounding results}
\begin{center}
\begin{tabular}{@{}ccccccccccccc@{}}
\toprule
       & \multicolumn{5}{c}{F1}         &\phantom{a} & \multicolumn{5}{c}{IoU}              & \multirow{2}{*}{ROC} \\ \cmidrule{2-6} \cmidrule{8-12} 
threshold & 0.15  & 0.20  & 0.25  & 0.30  & 0.35 && 0.15  & 0.20  & 0.25  & 0.30  & 0.35  &                      \\ \midrule
val    & .6950 & .7286 & 7412  & .7280 & .6700 && .5788 & .6328 & .6519 & .6254 & .5439 & .8857                \\
test   & .6953 & .7432 & .7626 & .7350 & .6380 && .5682 & .6296 & .6578 & .6203 & .5161 & .9186                \\
\bottomrule
\end{tabular}
\end{center}
\label{tab:ground_result}
\end{table*}

\subsection{Experiment Setup}

\textbf{Dataset.} We train and evaluate our model in our GIER dataset. The dataset is split into training, validation and testing subset with ratio 8:1:1, resulting in 4934/618/618 image pairs, respectively. \\

\noindent\textbf{Metrics.} For operation prediction, it is a multi-label classification task, hence we evaluate it using F1 score and ROC-AUC. For the operation conditioned grounding, we evaluate two sub-tasks: operation binary classification (local or global) evaluated by accuracy, and the local operation grounding evaluated by F1 score, ROC-AUC and IoU. Since the local operation grounding is formulated as a multi-object retrieving task, F1 score and ROC-AUC are reasonably set as the metrics. Moreover, the selected multiple objects can make up a whole image-level mask, so we also evaluated the mask quality using IoU score computed between the grounded mask and the ground truth mask.
To evaluate the final output image, we adopt L1 distance between the predicted image and the target image, where the pixel are normalized from 0 to 1. However, since the request could have many suitable editing, we further conduct human study to get more comprehensive evaluation. The implementation detail is in App.~\ref{sec:exp_detail}.

\begin{figure}[t]
    \centering
    \includegraphics[width=\columnwidth]{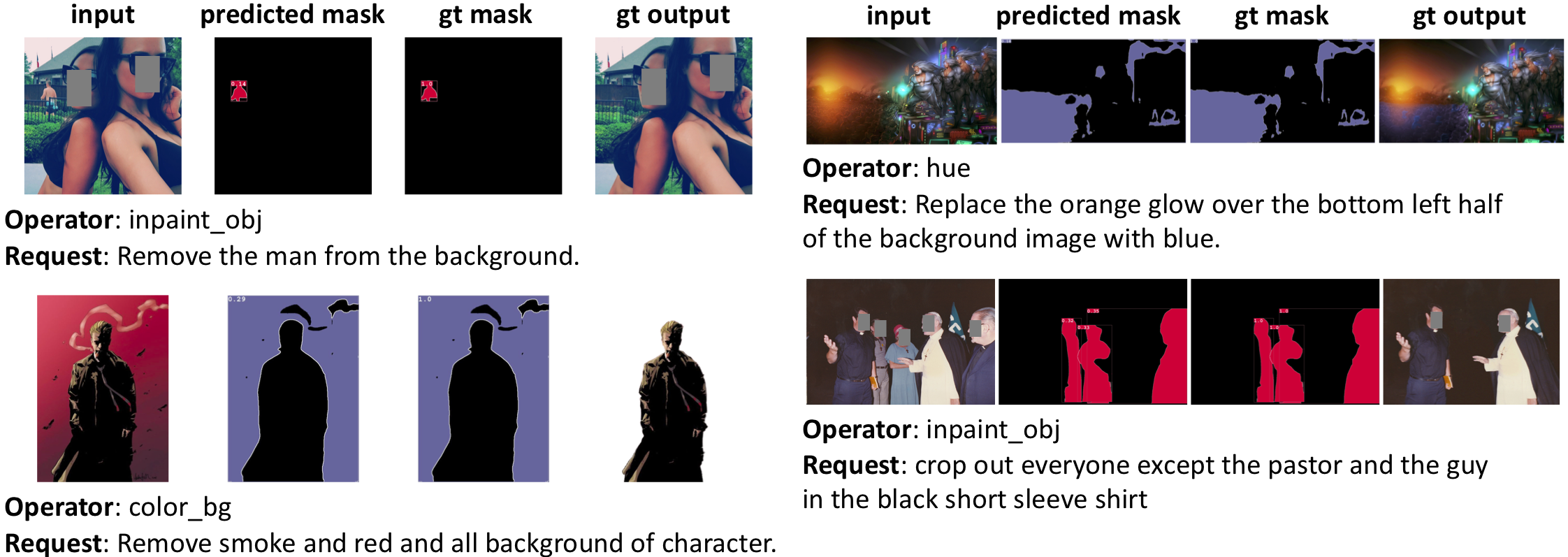}
    \caption{Visualization for the operation conditioned grounding}
    \label{fig:goodGround}
\end{figure}


\subsection{Results: Operation Prediction}
The result for operation prediction is shown in Tab.~\ref{tab:multilabel_f1}. We evaluate F1 score under different confidence thresholds and observe that the validation and test set has the similar trend and achieve best performance at threshold 0.4. And the ROC score also indicate a good performance on operation prediction and can support the later task well. Its visualization can be found in App.~\ref{sec:op_pred}.

\subsection{Results: Operation Conditioned Grounding}
For operation type classification, the accuracy is listed in Tab.~\ref{tab:operation_type}. 
For local operation grounding, the quantitative result is in Tab.~\ref{tab:ground_result}. F1 score and IoU are evaluated under various confidence thresholds with the same trend, and both attain peak value at threshold 0.25. The ROC score is 0.8857 and 0.9186 for validation and test set, respectively. The evaluation result indicating a good start for the operation modular network. Fig.~\ref{fig:goodGround} shows the qualitative grounding results for local operations. In many cases the request is to remove distraction persons in the background, such as the first and last row in Fig.~\ref{fig:goodGround}, requiring the grounding model to distinguish the high-level semantic of foreground and background. Also, the cartoon figures images make the grounding even more challenging. 
The visualization of the operation attention is in App.~\ref{sec:op_attn_vis}.
\subsection{Results: Language Driven Image Editing}
\label{sec:tmp}

 \begin{table}[t]
    \parbox{0.58\linewidth}{
    \begin{center}
    \caption{The comparison between a GAN-based method with our method. The arrow indicates the trend for better performance.}
    \scalebox{0.85}{
    \begin{tabular}{@{}lccc@{}}
    \toprule
     & L1$\downarrow$ & User Rating $\uparrow$ & User Interact $\uparrow$\\ \midrule
    Target & - & 3.60 & -\\
    Random Edit & 0.1639 & - & - \\
    Pix2pixAug~\cite{wang2018learning}& \textbf{0.1033} & 2.68 & 13.5\% \\ \midrule
    Our method (UB) & 0.0893  & - & - \\
    Our method& 0.1071 & \textbf{3.20} & \textbf{86.5\%}\\ \bottomrule
    \end{tabular}}
    \label{tab:gan_comp_exp}
    \end{center}}
    \hfill
    \parbox{0.38\linewidth}{
    \caption{Ablation study 1 and 2 with V and L representing vision and language}
    \begin{center}
        \begin{tabular}{@{}cccc@{}}\toprule
        Study    & Metric  & L & V+L \\ \midrule
        1& ROC     & 0.9182   & 0.9153          \\ \midrule
        \multirow{2}{*}{2} & ROC   & 0.9804    & 0.8969        \\ 
            & Acc@0.5 & 0.9508    & 0.9397       \\ \bottomrule
        \end{tabular}
    \end{center}
    \label{tab:abla_VL}}
\end{table}

\begin{figure}[t]
    \centering
    \includegraphics[width=0.85\columnwidth]{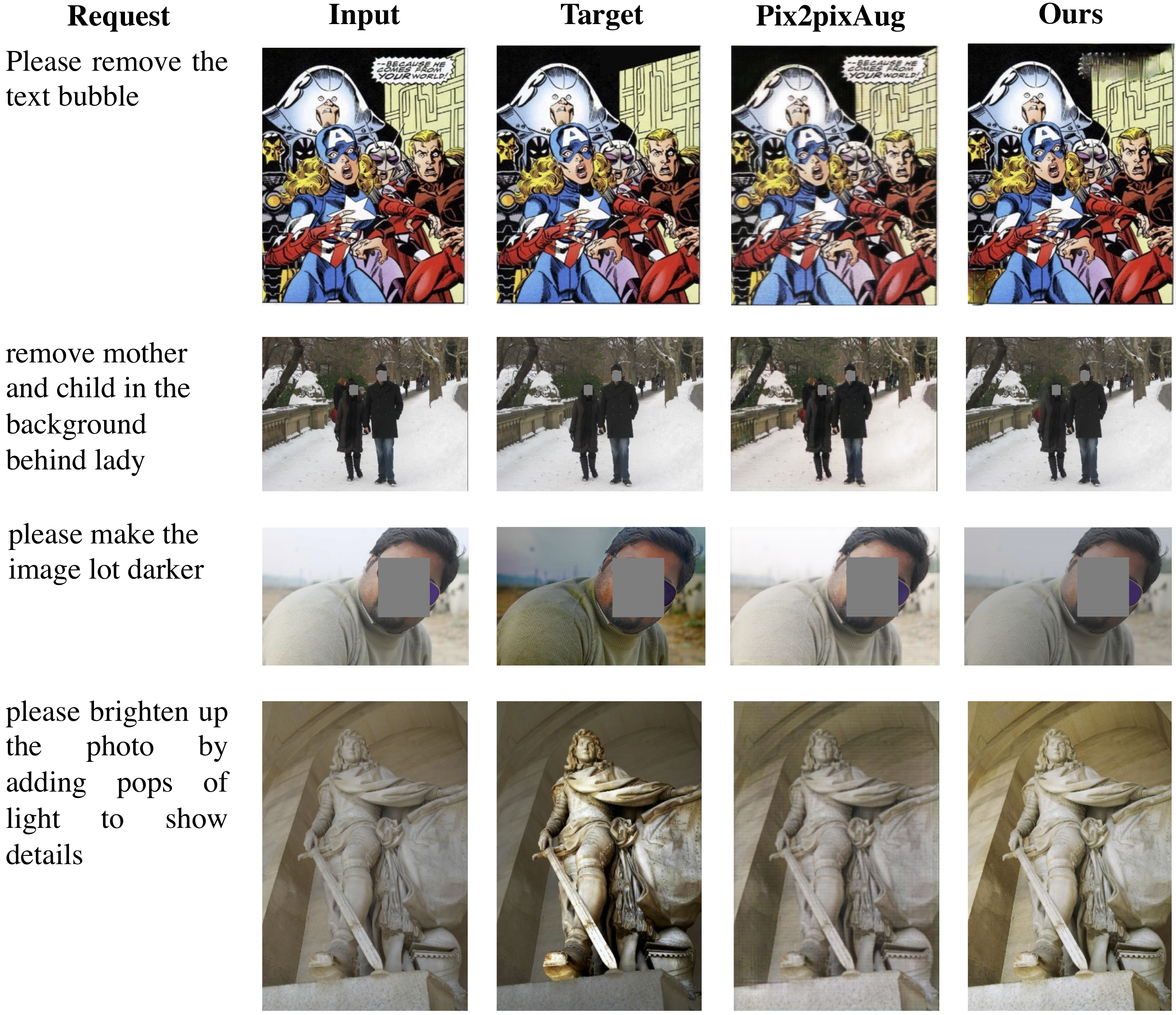}
    \caption{The visual comparison between our method and a GAN-based method. The first two rows are local editing, our method can correctly remove the designated object, even for text, while pix2pixAug cannot do such local editing. And for the last two rows, our method has more salient editing than pix2pixAug.}
    \label{fig:vis_comp_exp}
\end{figure}

The main quantitative results are shown in Tab.~\ref{tab:gan_comp_exp} with L1 and two user evaluation metrics. The comparison methods are described as follows.
\emph{Pix2pixAug} is a GAN-based model following the language-augmented pix2pix model in \cite{wang2018learning}.
\emph{Random Edit} is sequentially apply random editing operations with random parameters in random number of steps.
\emph{Our method UB} is the performance upper bound for OMN where the ground truth operations and masks are given as input.
\emph{Our method} is our full model where operations and masks are predicted.
Experiments show that Pix2pixAug has slightly better L1 score, but the user rating and user interactive ratio (detailed in App.~\ref{sec:usr_stdy}) strongly indicates that our method is more perceptually appealing to humans, and of more advantageous for human-interactive editing.
Also, the performance gap between our method and its upper bound indicates that better operation and mask prediction can bring a large performance gain.
Figure~\ref{fig:vis_comp_exp}  demonstrates that our method has better awareness for local editing and more salient editing effect than Pix2pixAug. More visualization of our edited images is in App.~\ref{sec:more_vis_res}.

\begin{table}[t]
    \parbox{0.49\linewidth}{
        \caption{The comparison between OMN with triplet loss and without triplet loss.}
        \begin{center}
        \begin{tabular}{@{}ccc@{}}
        \toprule
           & w/ Triplet & w/o Triplet \\ \midrule
        L1 & 0.0893           & 0.0925      \\ \bottomrule
        \end{tabular}
        \end{center}
        \label{tab:compare_triplet_loss}
    }
    \hfill
    \parbox{0.49\linewidth}{
        \caption{The comparison between OMN with fixed and random operation order.}
        \begin{center}
        \begin{tabular}{@{}ccc@{}}
        \toprule
        & Fixed order & Random order \\ \midrule
        L1 & 0.0893 & 0.0875 \\ \bottomrule
        \end{tabular}
        \end{center}
        \label{tab:order_ablation}
   }
\end{table}
\subsection{Ablation Study}
\textbf{Study 1:}\quad To investigate the importance of the visual information, we compare the operation prediction performance by using 1) only language feature (L) and 2) concatenation of vision and language feature (V+L). The result is listed in Tab.~\ref{tab:abla_VL}. We find that pure language feature is comparable with both vision and language feature, indicating that the language information itself usually already contains rich context for operation selection.

\noindent\textbf{Study 2:}\quad Also for the grounding task, we compare the global or local classification with or without visual information provided . The comparison is drawn in Tab.~\ref{tab:abla_VL}. It reveals that purely using language feature is a better way to decide if an operation is local or global. We suspect the reason is that if the operation is described with a location or object phrase, then such operation is of high possibility to be a local operation, so the visual information may not be so helpful compared with language.

\noindent\textbf{Study 3:}\quad we explore the effectiveness of the triplet loss applied on each generation step in upper bound setting.  
Table.~\ref{tab:compare_triplet_loss} shows that with the triplet loss the OMN achieves better performance, demonstrating its positive effect.

\noindent\textbf{Study 4:}\quad The effect of the operation order is evaluated in Tab.~\ref{tab:order_ablation} in upper bound setting. 
We compare the models trained and test in fixed order and random order, and the result is slightly better for random order than fixed order.

\section{Conclusion and Future Direction}
\label{sec:conclusion}
In this paper, we propose the LDIE task along with a new GIER dataset which supports both local and global editing and provides object masks and operation annotations. We design a baseline modular network to parse the whole request and execute the operation step-by-step, leading to an interpretable editing process. 
To handle the unique challenges of visual grounding in this new task, we propose the operation conditioned grounding model extending the MattNet to consider operation input and multi-region output. 

Currently our model uses the intermediate operation and mask as supervision to facilitate the modeling and in turn evaluate the annotation quality.
However such intermediate operation annotation might contain human bias and how to learn the model that only supervised by target image can be further explored. 
For evaluation metrics, LDIE task should also evaluate whether the edit is applied to the correct region specified by language. We evaluated this according to the grounding performance, which rely on the intermediate mask ground truth. However, a more general evaluation only depending on target image can be proposed.
Finally, more editing operations can be added to the model.

\noindent{\textbf{Acknowledgement}: This work was partly supported by Adobe Research, NSF 1741472 and 1813709. The article solely reflects the opinions and conclusions of its authors but not the funding agents.}


\bibliographystyle{splncs}
\bibliography{egbib}

\clearpage
\appendix
\begin{center}
\Large{\textbf{Appendix}} \\
\end{center}
In this appendix, we first introduce the user study process and  present more editing results of our model (App.~\ref{sec:comp_gan}). Then, we show more intermediate experimental results (App.~\ref{sec:exp_res}), describe the implementation of operations (App.~\ref{sec:op_design}) and experiment implementation details (App.~\ref{sec:exp_detail}). Finally, we elaborate the data collection interface (App.~\ref{sec:interface}) and visualize the two data collection criteria of our dataset (App.~\ref{sec:data_vis}).

\section{User Study and More Visualization Results}
\label{sec:comp_gan}
In this section, we firstly describe the user study and explain the related metrics. Then present more visualization of our editing methods.

 \begin{table}[]
    \begin{center}
    \caption{The comparison between a GAN-based method with our method. (This table is the same with Tab. \textcolor{red}{6} in main paper, shown here for better explaination of user evaluation.)}
    \begin{tabular}{@{}lccc@{}}
    \toprule
     & L1 & User Rating & User Interact \\ \midrule
    Target & - & 3.60 & -\\
    Random Edit & 0.1639 & - & - \\
    Pix2pixAug~\cite{wang2018learning}& \textbf{0.1033} & 2.68 & 13.5\% \\
    Our method& 0.1071 & \textbf{3.20} & \textbf{86.5\%}\\ \bottomrule
    \end{tabular}
    \label{tab:gan_comp_exp}
    \end{center}
\end{table}


\begin{figure*}[t]
    \centering
    \includegraphics[width=\columnwidth]{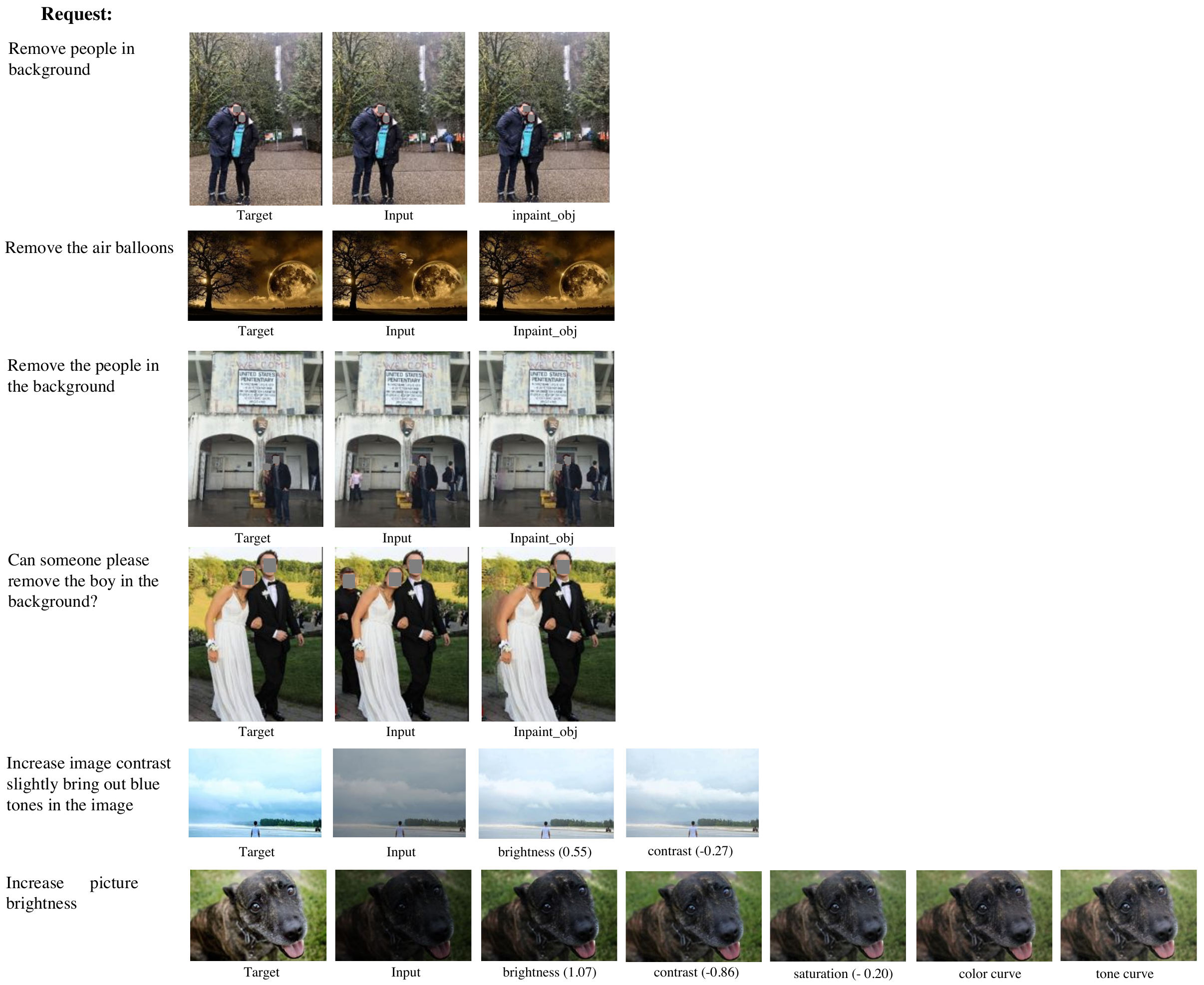}
    \caption{Visualization of the editing results including the intermediate steps from the whole pipeline of our method.}
    \label{fig:integ_vis_res}
\end{figure*}

\subsection{User Study Setup}
\label{sec:usr_stdy}
We conduct two user studies to evaluate the editing quality and the feasibility for human-interactive editing on Amazon Mechanic Turk (AMT).
For each user study, we random select 250 unique source images from the test set and each edit will be evaluated twice.
\\
\noindent\textbf{User Rating.} 
\textit{User rating} reflects the user perceptual evaluation for editing quality.
We collect user rating by showing users with input image, language request, and random order of target image, editing results of pix2pixAug and our method (user does not know which image corresponds to which method).
We let 38 users to rate score from 1 (worst) to 5 (best) for each of the edited image, and average over all users and images.\\
\noindent\textbf{User Interact.}
\textit{User Interact} indicates the feasibility of editing method for human-interactive editing.
We measure such feasibility by showing sequence of editing images from our method and the editing image from Pix2pixAug for users, and let users choose the image based on which they will follow up their own edits to complete the editing request. 
In total, 26 users are involved in this user study.
We show the percentage of each method that users would like to start with in Tab.~\ref{tab:gan_comp_exp}.
And it manifests 86.5\% chance that users prefer follow up editing based on the editing generated by our methods.
Among the users who choose our method, they prefer the image at 66.56\% of the total length of the editing sequence, indicating that the intermediate images are more favored for human-interactive editing. 

\subsection{More Visual Results}
\label{sec:more_vis_res}
Figure~\ref{fig:integ_vis_res} shows more our editing results.

\section{More Experiment Results}
\label{sec:exp_res} 
 \subsection{Operation prediction}
 \label{sec:op_pred}
\begin{figure}[]
    \centering
    \includegraphics[width=1\columnwidth]{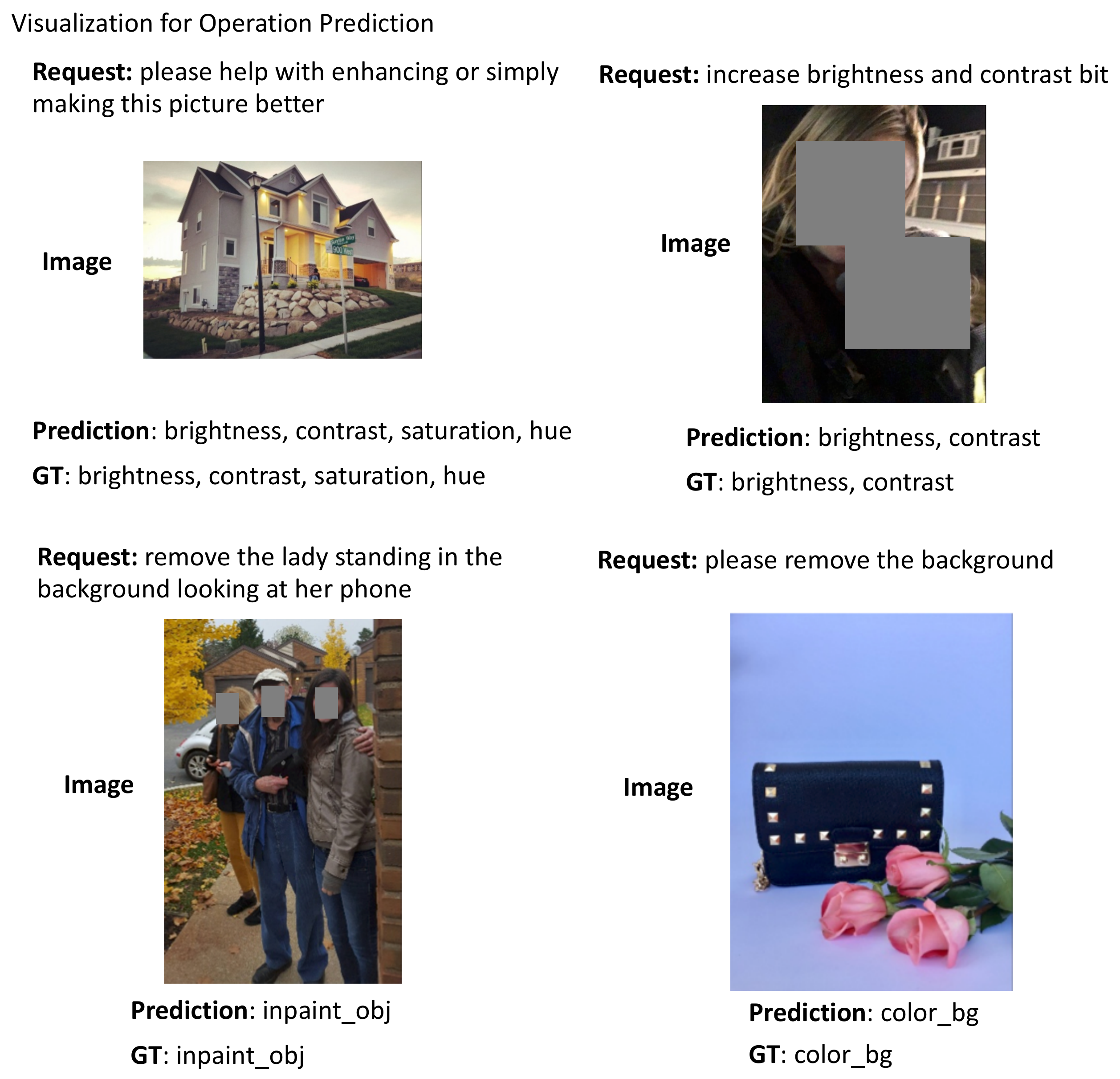}
    \caption{The visualization of operation prediction. The operation prediction model takes in the request and the input image, outputs the predicted operations. The ground truth operations (GT) are listed for reference.}
    \label{fig:op_pred_res}
\end{figure}

\begin{figure}[]
    \centering
    \includegraphics[width=1\columnwidth]{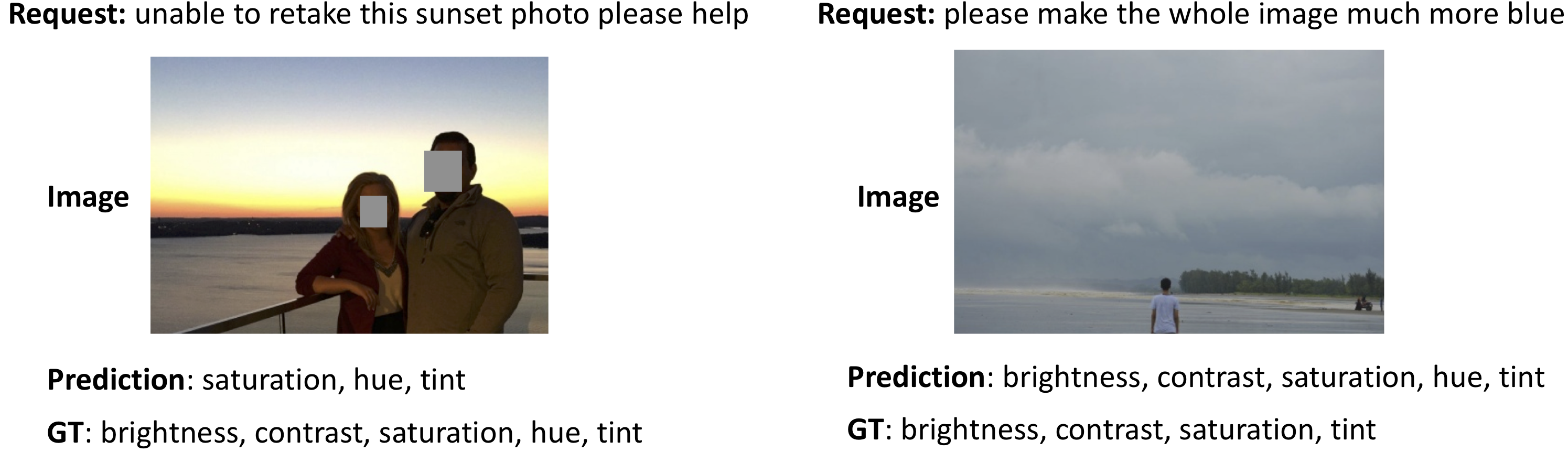}
    \caption{The failure cases for operation prediction.}
    \label{fig:op_pred_failure}
\end{figure}

The visualization of our predicted operation against ground truth is shown in Fig.~\ref{fig:op_pred_res}.
The failure case is shown in Fig.~\ref{fig:op_pred_failure}, indicating the model might miss or over predict the operations if multiple operations are applied.

\subsection{Operation Attentions Visualization}
\label{sec:op_attn_vis}
\begin{figure*}
    \centering
    \includegraphics[width=\columnwidth]{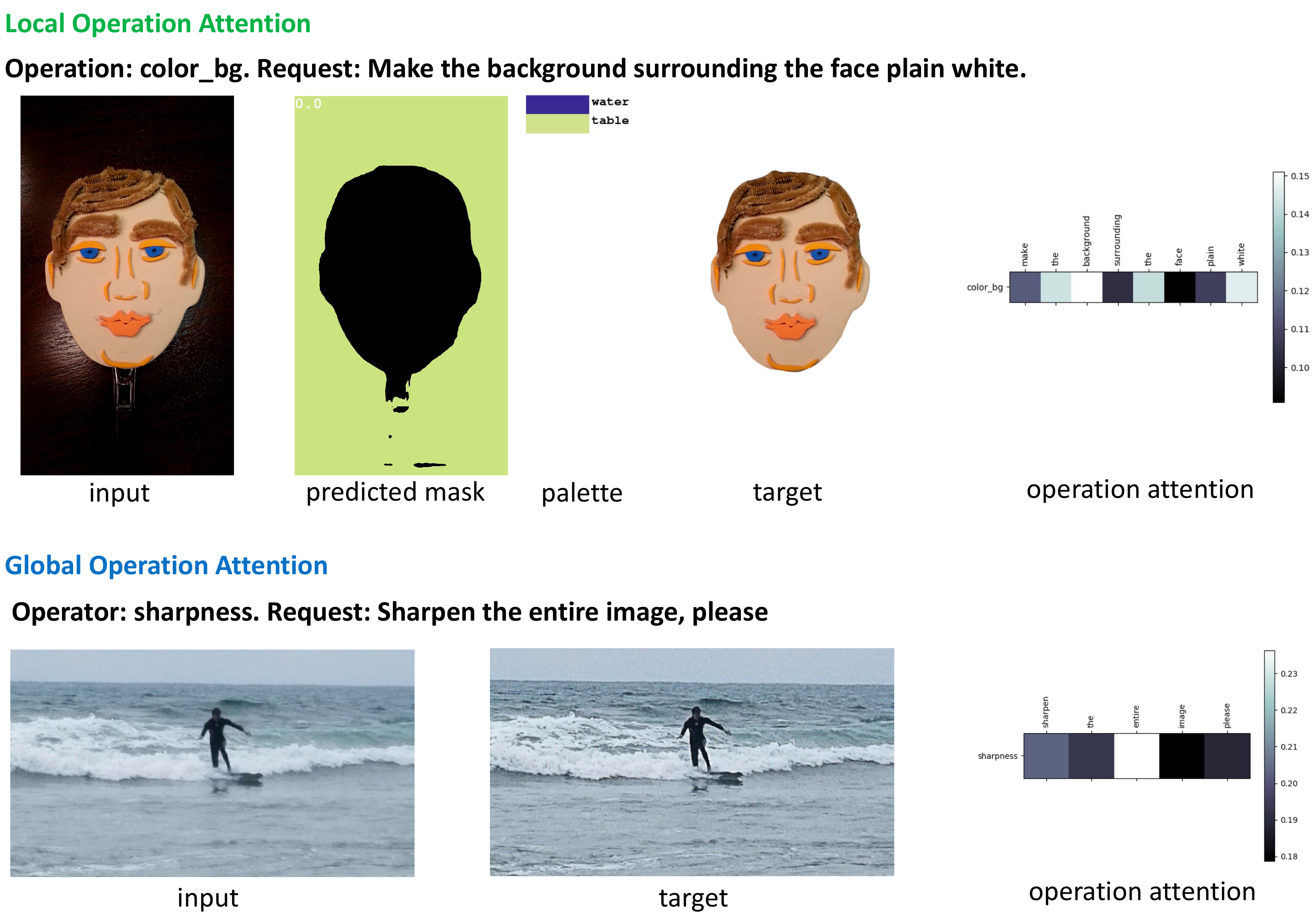}
    \caption{Example of the operation attention over the request. The left part is local operation, also showing the predicted masks.}
    \label{fig:op_attn}
\end{figure*}

The attention of the operation over the request is shown in Fig.~\ref{fig:op_attn}. The visual result shows the operation can attend to the key words indicating where the operation should be applied to.

\section{Operation Submodule Design}
\label{sec:op_design}
For the current task, we choose 8 operations: $\mathtt{brightness}$, $\mathtt{saturation}$, $\mathtt{contrast}$, $\mathtt{sharpness}$, $\mathtt{hue}$, $\mathtt{tint}$, $\mathtt{inpaint\_obj}$, $\mathtt{color\_bg}$. The operation modular network is composed of these operations in a fixed order if they are needed. With the input image $I$, parameter $p$, and output image $I'$,
the implementation of operation submodules are illustrated as follows.
\subsection{Brightness and Saturation}
The hue, saturation, value in the HSV space of image $I$ is denoted as $H(I)$, $S(I)$, $V(I)$.
Here $p$ is an unbounded scalar. Let $V'(I) = \text{clip}((1 + p)\cdot V(I), 0, 1)$ and $S'(I) = \text{clip}((1 + p)\cdot S(I), 0, 1)$, the output image for brightness operation is 
\begin{equation}
    I' = \text{HSVtoRGB}(H(I), S(I), V'(I)),
\end{equation}
and the output image for saturation operation is 
\begin{equation}
    I' = \text{HSVtoRGB}(H(I), S'(I), V(I)).
\end{equation}
The $\text{HSVtoRGB}$ is a differentiable function mapping the RGB space to HSV space, implemented via Kornia~\cite{eriba2019kornia}, and $\text{clip}(x, 0, 1)$ is a clip function to clip $x$ within $0$ to $1$.
\subsection{Contrast}
Contrast operation is controlled by a scalar parameter $p$, implemented following~\cite{hu2018exposure}.
First compute the luminance of image $I$ as 
\begin{equation}
    \text{Lum}(I) = 0.27 I_r + 0.67 I_g + 0.06 I_b,
\end{equation}
where $I_r$, $I_g$, $I_b$ are the RGB channels of $I$. 
The enhanced luminance is 
\begin{equation}
    \text{EnhancedLum}(I) = \frac{1}{2}(1 - \cos(\pi \cdot \text{Lum}(I))),
\end{equation}
and the image with enhanced contrast is 
\begin{equation}
    \text{EnhancedC}(I) = I\cdot \frac{\text{EnhancedLum}(I)}{\text{lum}(I)}.
\end{equation}
The output image $I'$ is the combination of the enhanced contrast and original image
\begin{equation}
    I' = (1-p)\cdot I + p\cdot \text{EnhancedC} (I)
\end{equation}
\subsection{Sharpness}
The sharpness operation is implemented by adding to the image with its second-order spatial gradient~\cite{gonzales2002digital}, expressed as 
\begin{equation}
    I' = I + p\Delta^2I,
\end{equation}
where $p$ is a scalar parameter and $(\Delta^2\cdot)$ is the Laplace operator over the spatial domain of the image. The Laplace operator is applied to each channel of the image.
\subsection{Tint and Hue}
The tint and hue operation follows curve representation~\cite{hu2018exposure}. The curve is estimated as piece-wise linear functions with $N$ pieces. The parameter $p=\{p_i\}_{i=0}^{M-1}$ is a vector of length $M$. With the input pixel $x\in [0, 1]$, the output pixel intensity is 
\begin{equation}
    f(x) = \frac{1}{Z}\sum_{i=0}^{N-1}\text{clip}(N x-i, 0, 1)p_i,
\end{equation}
where $Z=\sum_{i=0}^{N-1}p_i$. 
For tint operation, $N=M=8$, the same $f$ will apply to each of the RGB channels of the image $I$.
For hue operation, three different $f$ are applied individually to each of RGB channels. Each $f(x)$ has $N=8$, leading to $M=3N=24$.

\subsection{Inpaint\_obj and Color\_bg}
$\mathtt{inpaint\_obj}$ indicates inpainting object and $\mathtt{inpaint\_obj}$ denotes color background. The inpainting is implemented by EdgeConnect~\cite{nazeri2019edgeconnect}.
Since the $\mathtt{color\_bg}$ operation plays the major role at removing the background, so we force the color to be white, so that the operation will be applied to the background mask and make the background white.
$\mathtt{inpaint\_obj}$ is also not resolution-independent because it is implemented using a neural network.
Fortunately, these two operations do not need parameters, so we put them at the first positions in the modular network, allowing the following operation modules fully differentiable w.r.t the final output image.

\section{Experiment Details}
\label{sec:exp_detail}
The image is normalized from zero to one.
Training images are resized to $128\times 128$, and testing images are resized to short side $300$ but capped long side $500$, keeping aspect ratio unchanged.
For operation prediction and operation conditioned grounding, they are optimized with Adam with initial learning rate 0.0004 and batch size 16. 
The learning rate is decreased by a factor of 10 every 8000 iterations after the first 8000-iteration warm-up. 
The word embedding size, operation embedding size and hidden size for LSTM are all 512, and all MLP and FC layers are 512-dimensional.
The image feature is extracted via the Global Average Pooling (GAP) layer after the last feature map of ResNet18, with dimension 512. 
In operation conditioned grounding, the model configuration for MattNet variant remains the same as MattNet. The binary cross-entropy loss for local/global classification and the ranking loss of MattNet has equal balance weight. The MattNet variant does not start to train until finishing 1000 iterations of warm-up for the local/global classifier. The mask proposal feature for MattNet variant is extracted by applying GAP to the feature map masked with each mask proposal, where the feature map is the output of the semantic head in UPSNet~\cite{xiong2019upsnet}, and the mask proposal is the predicted panoptic mask by UPSNet.

For operation modular network, it is optimized with Adam with learning rate 0.00001. The LSTM has two hidden layers stacked with 512 dimension each. The word embedding and operation embedding are of dimension 300 where the word embedding is initialized with Glove feature~\cite{pennington2014glove}. The MLP in each operation submodule is equipped with batch norm at the final FC layer to prevent over-fitting.
The balance weight $\lambda$ is set $1$. 
Operation prediction, operation conditioned grounding, and operation modular network are trained in 20k iterations with best model at 12k, 14k, 16k iteration, respectively.
And for integral testing, the confidence thresholds for operation prediction, local/global classification, and grounding matching score are set $0.4$, $0.5$, and $0.25$, respectively.

\section{Data Collection Interface}
\label{sec:interface}
\begin{figure}
    \centering
    \includegraphics[width=0.7\columnwidth]{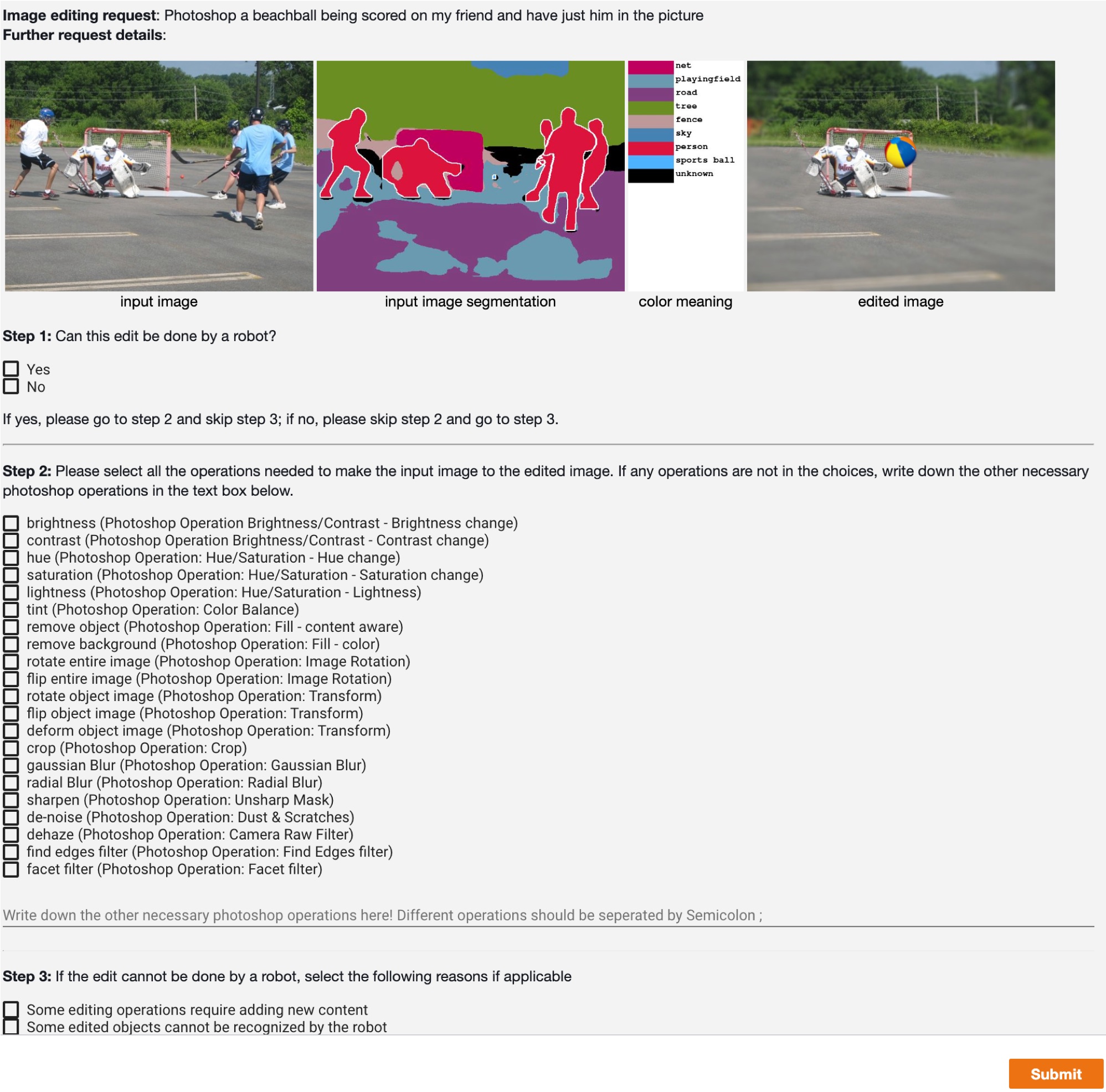}
    \caption{The interface for the editing validity checking according to the two criteria and the selection of feasible operations.}
    \label{fig:interface_round_1}
\end{figure}

\begin{figure}
    \centering
    \includegraphics[width=0.7\columnwidth]{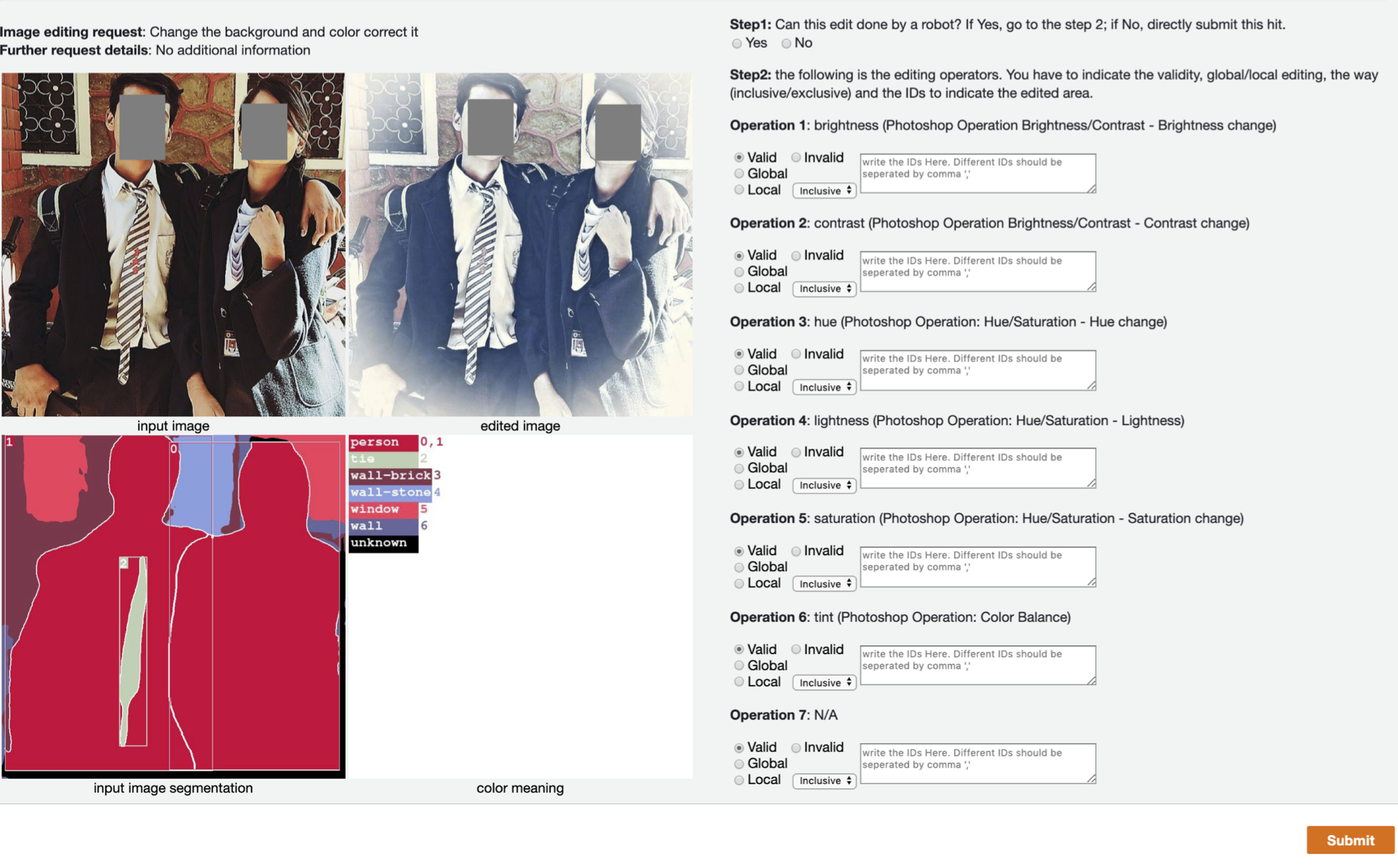}
    \caption{The interface for annotating the masks corresponding to operations. The input image segmentation is the panoptic segmentation with each segment tagged a unique ID. 
    Workers will first check whether each operation is valid. If valid, they further annotate whether the operation is global or local. If local, they finally should write down IDs of the segments to which the operation is applied to.}
    \label{fig:interface_round_2}
\end{figure}

\begin{figure}
    \centering
    \includegraphics[width=0.7\columnwidth]{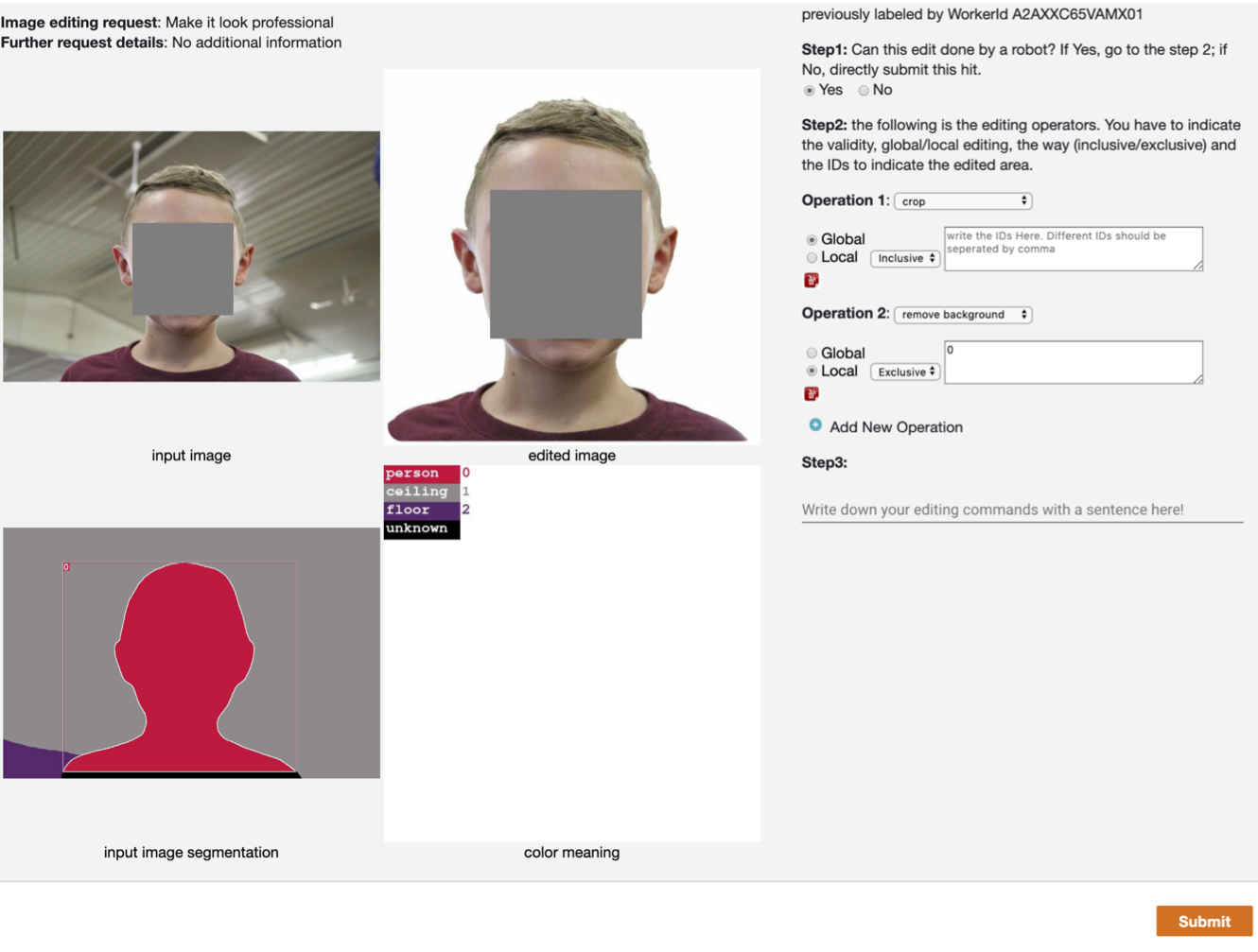}
    \caption{The interface for expert workers to label the language request and check the previously annotated operations and masks. The worker have to check the previously annotated operations and masks and overwrite the incorrect annotations with correct ones, and also write a professional language request.}
    \label{fig:interface_round_3_upwork}
\end{figure}

\begin{figure}
    \centering
    \includegraphics[width=0.7\columnwidth]{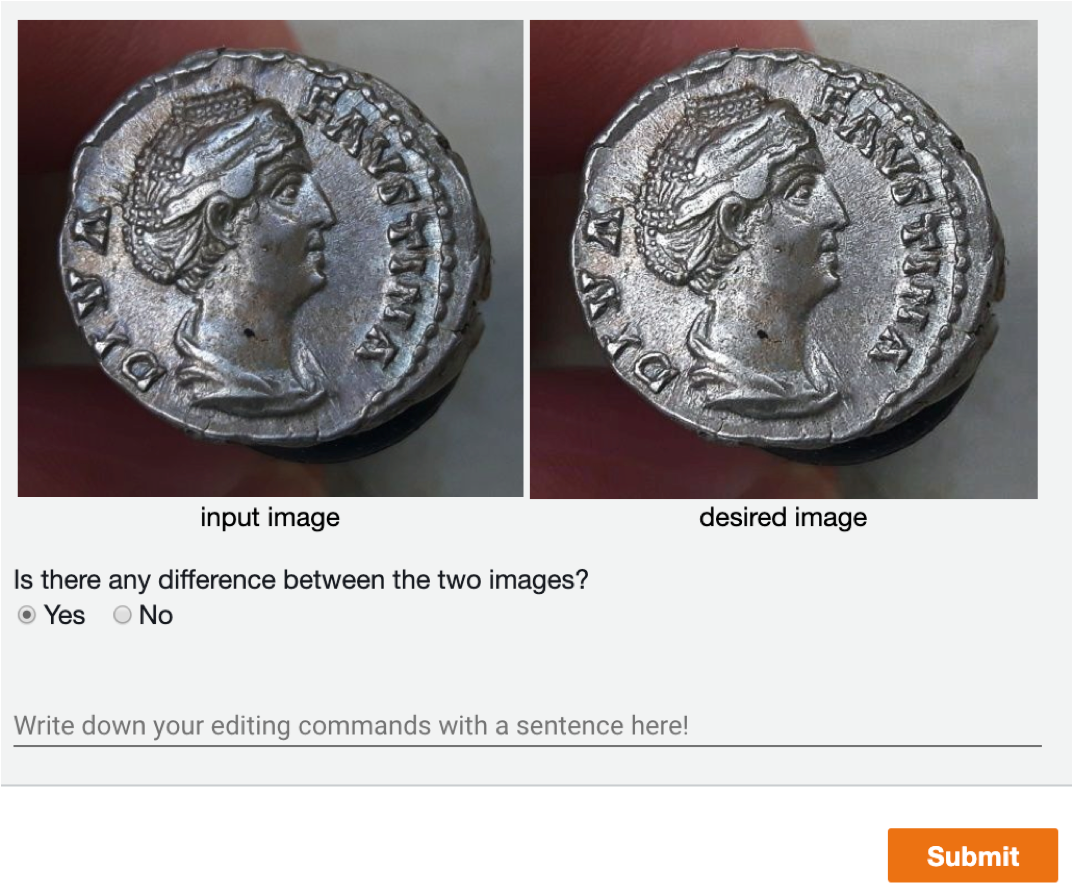}
    \caption{The interface for amateur workers to label the language request.}
    \label{fig:interface_round_3_AMT}
\end{figure}

\begin{figure}
    \centering
    \includegraphics[width=0.7\columnwidth]{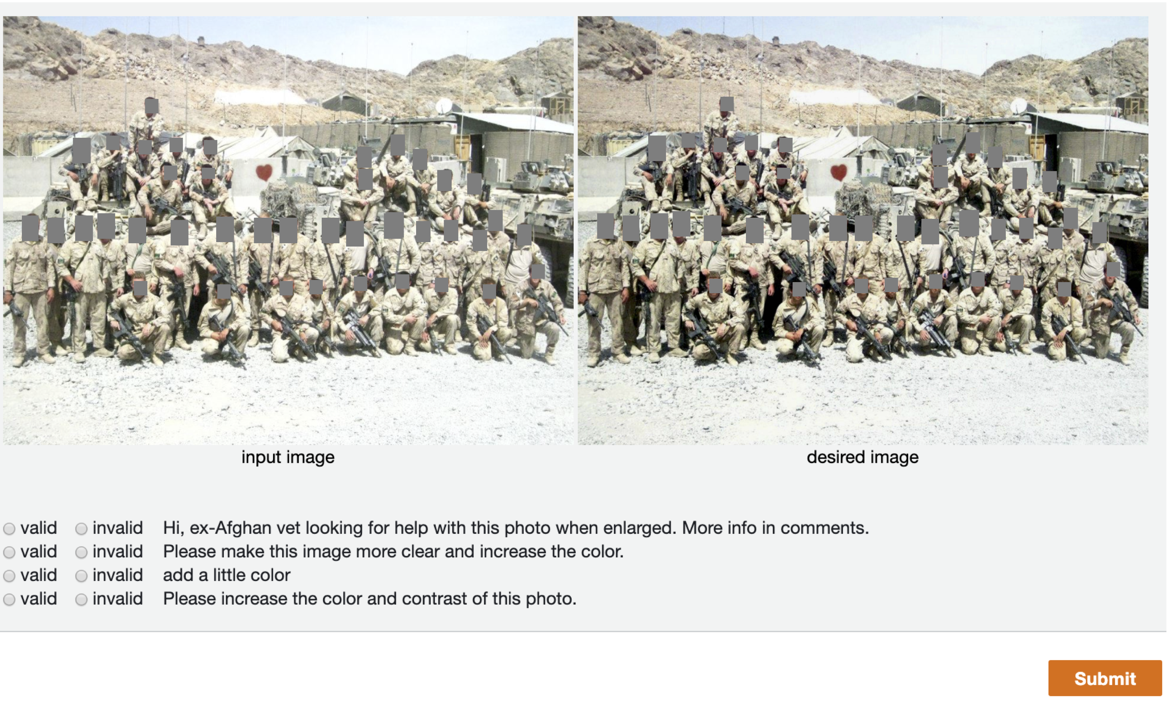}
    \caption{The interface for checking the quality of amateur-annotated request, with mixture of the original crawled language request.}
    \label{fig:interface_round_4}
\end{figure}

Firstly, Fig.~\ref{fig:interface_round_1} shows the interface enabling workers to check the editing validity according to the two criteria (1. no new things or stuffs; 2. no edit to unknown region) and select feasible editing operations. All known regions are visualized in the image segmentation. 
Secondly, Fig.~\ref{fig:interface_round_2} is the interface to check the previous annotated operations and also collect the operation type (local or global) and the masks where operations are applied to.
Next, the language request annotation is collected through interface in Fig.~\ref{fig:interface_round_3_upwork} and Fig.~\ref{fig:interface_round_3_AMT}.
Figure~\ref{fig:interface_round_3_upwork} is the interface for expert workers, where they can see all the annotated operations and masks and write profession-styled requests. Meanwhile, they should also check the correctness of previously annotated operations and masks and overwrite the incorrect ones.
Figure~\ref{fig:interface_round_3_AMT} is the interface for amateur workers, where they can only see a image pair and are required to write a request.
For the quality control, an extra interface shown in Fig.~\ref{fig:interface_round_4} is designed to check the amateur-annotated request.

\section{Visualization of Two Data Collection Criteria}
\label{sec:data_vis}
\begin{figure}
    \centering
    \includegraphics[width=0.6\columnwidth]{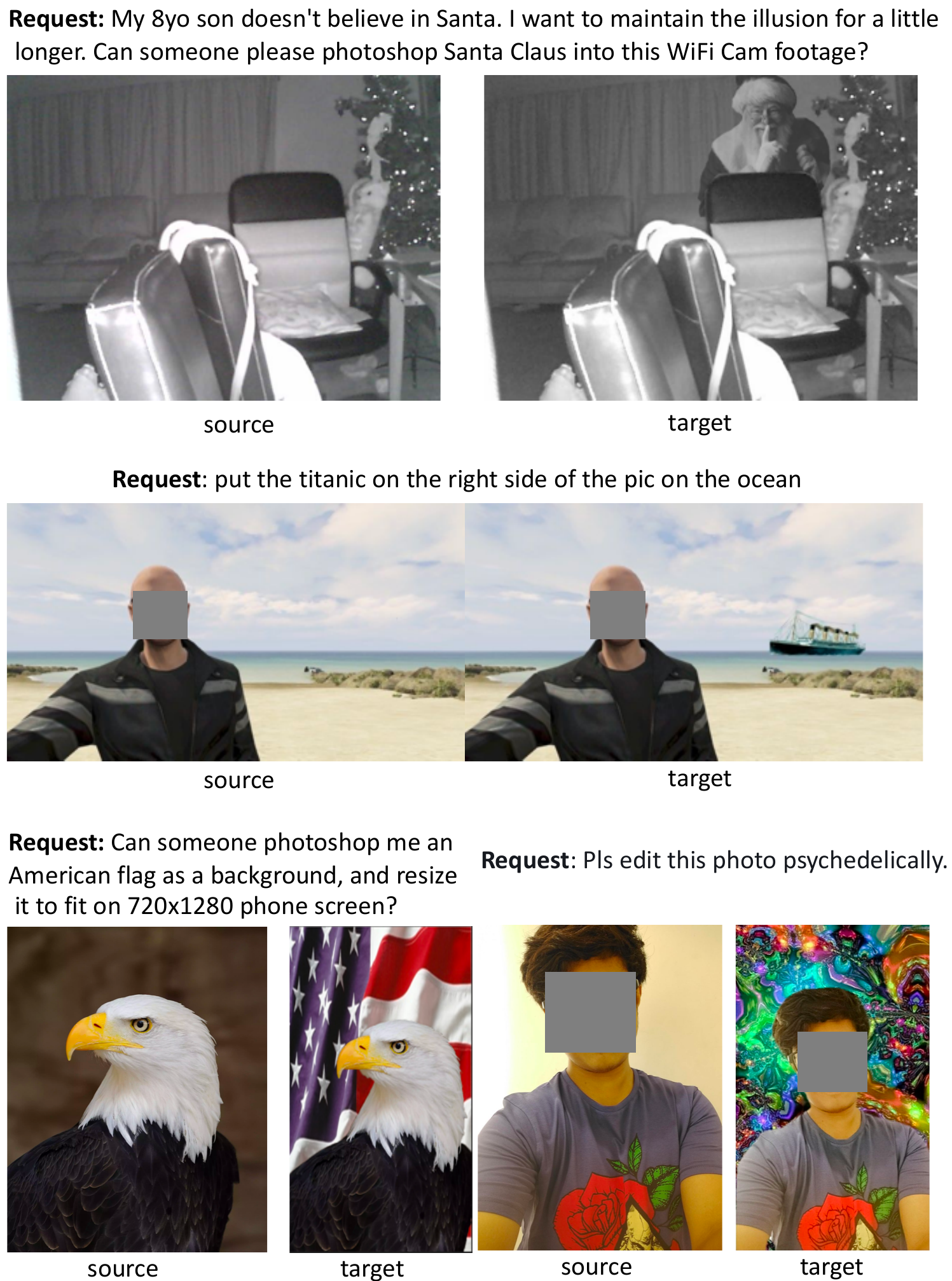}
    \caption{The crawled data triplets (source image, target image and request) where novel things or staffs are added. The first two rows contain novel objects, and the last row contains new backgrounds}
    \label{fig:dataset_novel_obj}
\end{figure}

\begin{figure}
    \centering
    \includegraphics[width=0.7\columnwidth]{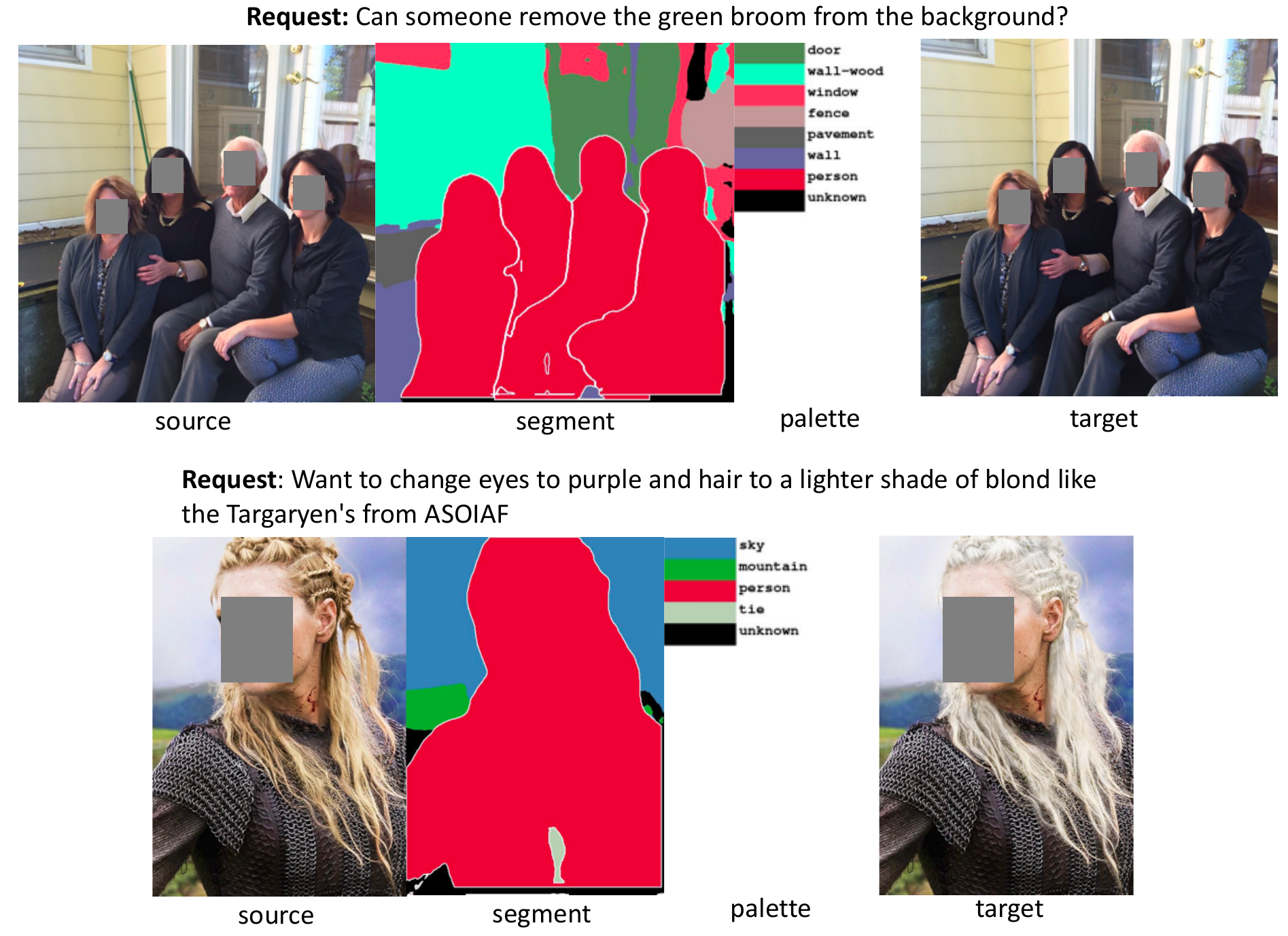}
    \caption{The crawled data samples where objects cannot be localized by our grounding model. The segmentation is obtained from UPSNet and the palette indicates the categories for each color.}
    \label{fig:dataset_unknown_obj}
\end{figure}

The collected image are extremely diversified with two major challenges: 1) edited images contains novel thing such as `add a frame to the image', and are struggling to learn. 2) some edited area is hard to localize. Thus two criteria to get a simpler starting point: 1) the edited image should not have new thing added, for example, Fig.~\ref{fig:dataset_novel_obj}; 2) the edited region can be localized by our grounding model, e.g., Fig.~\ref{fig:dataset_unknown_obj} . We let Photoshop experts to filter out those images violating the two criteria, by presenting them with the image pairs, editing requests, and the image panoptic segmentations. 

\end{document}